\documentclass{article}

\usepackage[preprint]{neurips_2026}

\usepackage[utf8]{inputenc} 
\usepackage[T1]{fontenc}    
\usepackage{hyperref}       
\usepackage{url}            
\usepackage{booktabs}       
\usepackage{amsfonts}       
\usepackage{nicefrac}       
\usepackage{microtype}      
\usepackage{xcolor}         
\usepackage{amsmath} 
\usepackage{makecell}
\usepackage{graphicx}
\usepackage{tabularx}
\usepackage{booktabs}
\usepackage[most]{tcolorbox} 
\usepackage{fvextra} 
\tcbuselibrary{breakable,listings}
\usepackage[T1]{fontenc}
\usepackage{lmodern}

\title{When Text Misleads: Inconsistent-Aware Reasoning for Audio-Grounded Dialogue}

\author{Yen-Ju Lu$^{*}$, Yuzhe Wang$^{*}$, Yaohan Guan, Xiluo He, Jiarui Hai, \\ 
\textbf{Mingrui Liang, Kaavya Chaparala, Thomas Thebaud,} \\ \textbf{Laureano Moro-Velazquez, Najim Dehak, Jesus Villalba } \\ 
\\
Center for Language and Speech Processing, Johns Hopkins University}

\begin{document}

\maketitle

\begin{abstract}
Understanding spoken dialogue requires joint reasoning over lexical content and paralinguistic acoustic signals such as emotion and conversational intent. However, existing evaluations often allow shortcuts based on transcripts or single-modality solutions, obscuring whether models genuinely ground predictions in speech. We formalize this failure mode as cross-modal disagreement, where transcripts suggest plausible but incorrect surface interpretations while acoustic cues such as prosody or speaking style support different answers. We develop a scalable framework that identifies text-biased surface interpretations and converts disagreement regions into conflict QA examples. We also include consistent cases where transcript-based and speech-grounded interpretations agree, enabling evaluation beyond adversarial audio dependence. This results in \textsc{ContraTalk}, a controlled benchmark containing 501 questions across five discourse dimensions: interaction behavior, emotion state, dialogue act, social stance, and conversational intent. We further develop an agentic-style reasoning framework that converts speech into an \emph{Audio Twin}, a text-readable representation of localized acoustic cues that exposes acoustic evidence to the reasoning model. Experiments show that strong text-only LLMs exceed 90\% accuracy in consistent cases but drop to 33--48\% in conflict cases. Direct AudioLLMs provide only partial grounding, still selecting the transcript-biased trap in roughly 30--40\% of conflict cases. Our Audio Twin framework improves conflict-case accuracy while reducing trap selection, but its consistent-case behavior remains backbone-dependent. These results identify transcript-based shortcuts as an important failure mode in spoken dialogue understanding and show that explicit acoustic evidence aggregation provides a more controllable interface for diagnosing and improving speech-grounded reasoning.

\end{abstract}

\section{Introduction}
\begin{figure}[t]
    \centering
    \includegraphics[width=0.8\linewidth]{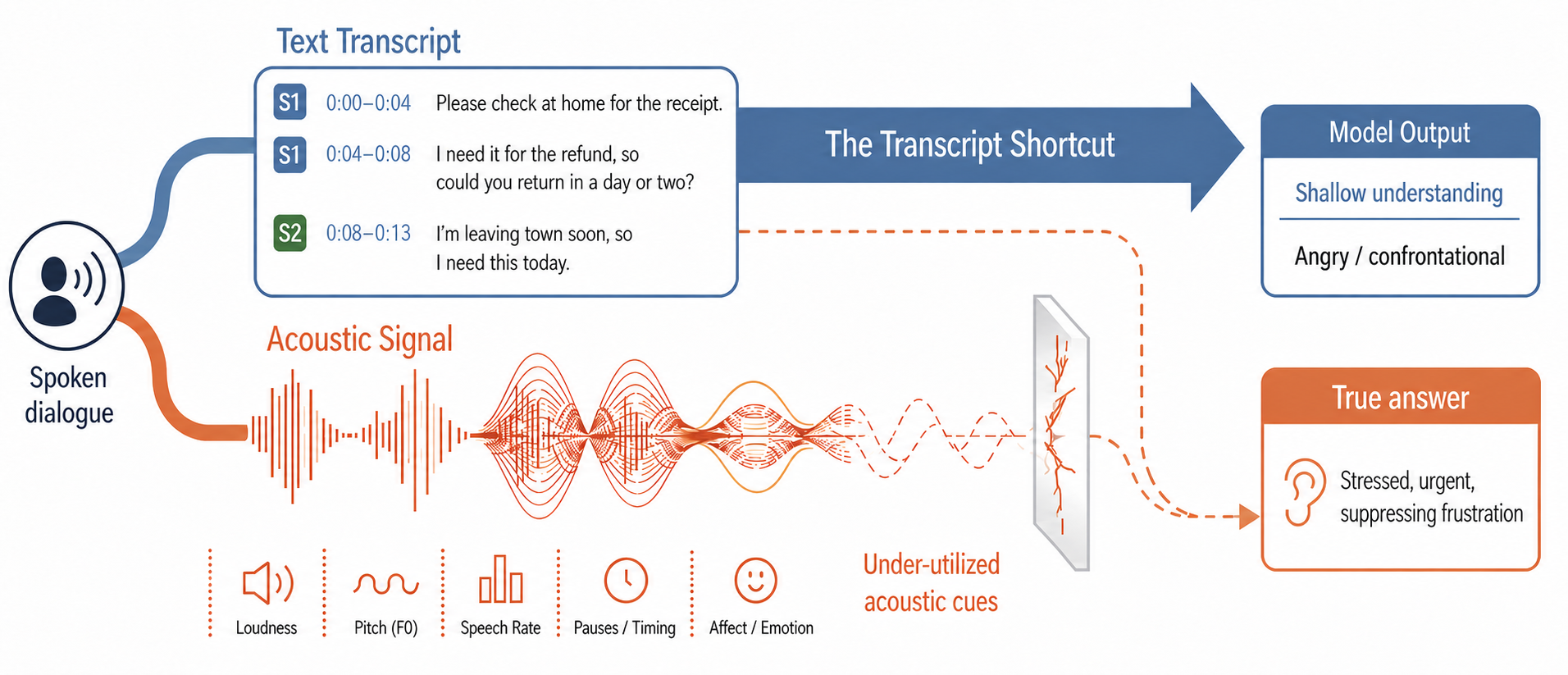}
    \caption{Existing spoken dialogue benchmarks often allow models to rely on transcript shortcuts, under-utilizing acoustic evidence that carries crucial paralinguistic information.}
    \label{fig:intro_shortcut}
    \vspace{-6mm}
\end{figure}


\begingroup
\renewcommand\thefootnote{}
\footnotetext{$^*$ denotes equal contribution.}
\addtocounter{footnote}{-1}
\endgroup

Multimodal understanding is often framed as a problem of integrating complementary signals. In spoken dialogue, transcripts convey lexical and semantic cues, while acoustic cues from the audio reveal interaction behavior, emotion state, social stance, and conversational intent. Yet these modalities do not always align: speakers may sound hesitant while making confident statements, suppress frustration beneath polite wording, or signal intent through timing and delivery patterns that are not recoverable from text alone. Such cases introduce a distinct reasoning challenge: resolving not only what is said, but how it is said.

Despite rapid progress in multimodal modeling, many evaluation settings still permit modality shortcuts. 
This issue has been widely observed in visual question answering, where models exploit language priors rather than genuine visual grounding \citep{goyal2017making}, and in conversational emotion recognition, where textual features often dominate acoustic cues \citep{poria2019emotion}. 
Recent audio-language studies similarly show that strong performance can arise from lexical shortcuts or dataset-specific priors without genuine acoustic grounding \citep[e.g.,][]{yang2024air, huang2024dynamic}. 
More generally, when a benchmark can be solved from only one modality, the other modality becomes functionally ignorable. 
This creates a form of \emph{modality collapse}: models may rely on transcripts when audio is needed, or rely on isolated audio cues without reasoning over dialogue context. 
A meaningful benchmark for spoken multimodal reasoning should therefore test whether models can compare, reconcile, and correct interpretations across modalities. 
We argue that cross-modal disagreement provides a distinct and underexplored axis for such evaluation.

To operationalize this perspective, we develop a scalable benchmark construction framework that starts from text-only surface interpretations, identifies regions where transcript-based and speech-grounded interpretations diverge, and converts them into controlled question-answering instances. 
The resulting benchmark, \textsc{ContraTalk}, evaluates grounded spoken dialogue reasoning across five discourse-level dimensions: interaction behavior, emotion state, social stance, dialogue act, and conversational intent. 
Rather than focusing only on conflict cases, \textsc{ContraTalk} also includes consistent cases where transcript-based and speech-grounded interpretations agree, enabling controlled evaluation of whether models use the appropriate evidence rather than collapsing to a single dominant modality.

Resolving such disagreement requires more than implicit multimodal fusion. 
A model must determine whether transcript-based reasoning is sufficient, identify which acoustic evidence is relevant, and compare competing interpretations when text and audio diverge. 
To support this process, we introduce \emph{Audio Twin}, a structured, text-readable representation of localized acoustic and interactional evidence. 
Audio Twin converts speech-derived cues such as prosody, affect, timing, and speaking behavior into evidence that can be retrieved and compared with transcript-based interpretations during reasoning. 
This provides an explicit interface for spoken dialogue understanding: rather than treating audio as opaque input, the reasoning model receives speech-grounded evidence in a form that can be inspected and used to revise or confirm the surface interpretation suggested by the transcript.

Experiments show that strong text-only LLMs perform well on consistent cases but degrade substantially under cross-modal disagreement, revealing the limits of transcript-centric reasoning. 
Direct Audio-LLMs reduce some transcript-biased errors but still frequently select text-biased traps, indicating that audio access alone does not ensure speech grounding. 
Our Audio Twin framework further improves conflict-case accuracy and reduces trap selection, suggesting that explicit acoustic evidence aggregation can provide a more controllable interface for spoken dialogue reasoning.

Our contributions are threefold:
\begin{itemize}
    \item We identify transcript-based shortcuts under cross-modal disagreement as an important failure mode in spoken dialogue understanding, and formalize this setting as requiring models to reconcile lexical and paralinguistic evidence.

\item We introduce \textsc{ContraTalk}, a controlled spoken dialogue QA benchmark constructed from text--audio disagreement regions, together with consistent cases that test whether models preserve transcript-supported reasoning when modalities agree.

\item We develop an agentic-style Audio Twin reasoning framework that converts localized speech-derived cues into text-readable evidence for reconciling transcript-based and speech-grounded interpretations.
\end{itemize}

\section{Related Work}
\noindent\textbf{Modality Bias and Shortcut Learning.}
Multimodal learning has long been challenged by representation imbalance, where models disproportionately rely on a dominant modality while underutilizing others \citep{wang2020makes}. 
This phenomenon is closely related to shortcut learning \citep{geirhos2020shortcut}, in which neural networks exploit easy-to-learn priors rather than integrating complex multimodal evidence. 
In spoken dialogue and emotion recognition, modality bias is particularly pronounced: textual transcripts often provide highly predictive semantic shortcuts, causing models to favor lexical cues while marginalizing paralinguistic and acoustic signals \citep{poria2019emotion, wagner2023dawn}. 
This imbalance is especially consequential in conversational settings, where meaning frequently emerges from pragmatic and affective cues expressed through prosody, timing, and interaction dynamics rather than lexical content alone. 
Although prior work has proposed debiasing strategies and regularization techniques \citep{wu2022characterizing}, existing evaluation paradigms rarely place models in scenarios where resolving explicit cross-modal disagreement is necessary for successful reasoning.

\noindent\textbf{Multimodal Reasoning and Evaluation for Audio-LLMs.}
The rapid development of Audio-LLMs has shifted spoken dialogue understanding toward unified end-to-end modeling paradigms, increasing the need for rigorous evaluation of multimodal reasoning capabilities \citep{tang2024salmonn, chu2023qwen}. 
In parallel, the community has introduced diverse benchmarks for audio-language understanding, compositional reasoning, instruction following, and conversational interaction, including AIR-Bench, AudioBench, and MMAU \citep{yang2024air, wang2025audiobench, sakshi2025mmau}.
Prior work has also explored modality-interaction phenomena such as multimodal sarcasm \citep{castro2019towards} and semantic meaning shifts in image--text pairs \citep{kiela2020hateful}. 
However, these benchmarks largely evaluate settings where modalities are cooperative, complementary, or independently informative. 
In contrast, \textsc{ContraTalk} targets cross-modal disagreement in spoken dialogue, where transcript-based interpretations can be plausible but wrong, and acoustic or interactional evidence is required to correct, refine, or disambiguate them.

\noindent\textbf{Agentic-style Multimodal Reasoning}
To handle complex multimodal inputs, recent approaches have adopted reasoning-and-acting paradigms that decompose problems and leverage external tools \citep{yao2023react, schick2023toolformer}. 
This trend has led to multimodal orchestration frameworks in which a central language model dynamically routes queries to specialized perceptual modules or engages in iterative retrieval loops to gather relevant context \citep{shen2023hugginggpt, huang2024audiogpt}. 
More recent work also uses structured digital-twin representations to expose perceptual states to LLM-based reasoning systems; for example, \citet{shen2025operating} construct an operating-room digital twin that preserves semantic and spatial relationships and supports a reason-retrieve-synthesize workflow for reasoning segmentation. 
Our Audio Twin follows this broader direction of making perceptual evidence explicit and LLM-readable, but shifts the focus from complementary evidence retrieval to cross-modal evidence arbitration. 
Rather than assuming retrieved evidence is mutually consistent with the text query, our framework explicitly resolves epistemic disagreement between transcript-derived interpretations and acoustic signals.

\section{Inconsistent-Aware Multimodal Reasoning}
\label{sec:problem_formulation}

\subsection{Problem Formulation}

We study spoken dialogue understanding as inference over paired textual and acoustic evidence.
A dialogue instance is represented as
$D=(T,A),$
where \(T\) denotes the transcript and \(A\) denotes the corresponding audio. The transcript primarily conveys lexical content, while the acoustic signal provides paralinguistic and interactional cues such as prosody and affect.
We define \emph{Cross-modal disagreement reasoning} as the setting where models must explicitly detect, interpret, and resolve conflicting evidence across modalities, rather than assuming alignment or relying on a single dominant modality. 
%
The goal is to evaluate whether models can use audio when it materially affects interpretation while remaining calibrated when the transcript and audio are aligned.
%
We formulate the task as multiple-choice question answering over spoken dialogue. 
Given a dialogue instance \((T,A)\), a question \(q\), and answer candidates \(\mathcal{C}=\{c_1,\dots,c_K\}\), we distinguish between two inference views:
\[
\hat{y}_{T}=f_T(T,q,\mathcal{C}), \qquad
\hat{y}_{TA}=f_{TA}(T,A,q,\mathcal{C}).
\]
Here, \(\hat{y}_{T}\) denotes the answer induced by transcript-only reasoning, and \(\hat{y}_{TA}\) denotes the answer supported by the full spoken dialogue, including acoustic evidence. 
We define two evaluation regimes:
\[
\text{conflict: } \hat{y}_T \neq \hat{y}_{TA},\; y^\star=\hat{y}_{TA},
\qquad
\text{consistent: } \hat{y}_T = \hat{y}_{TA} = y^\star .
\]
In conflict instances, the transcript induces a plausible but incorrect surface answer, while acoustic evidence determines the gold answer. 
In consistent instances, both modalities support the same answer, testing whether models remain calibrated when no cross-modal correction is required. 
Each question targets a discourse-level interpretation, such as interaction behavior, emotion state, social stance, dialogue act, or conversational intent.


\begin{figure}[t]
    \centering
    \includegraphics[width=\textwidth]{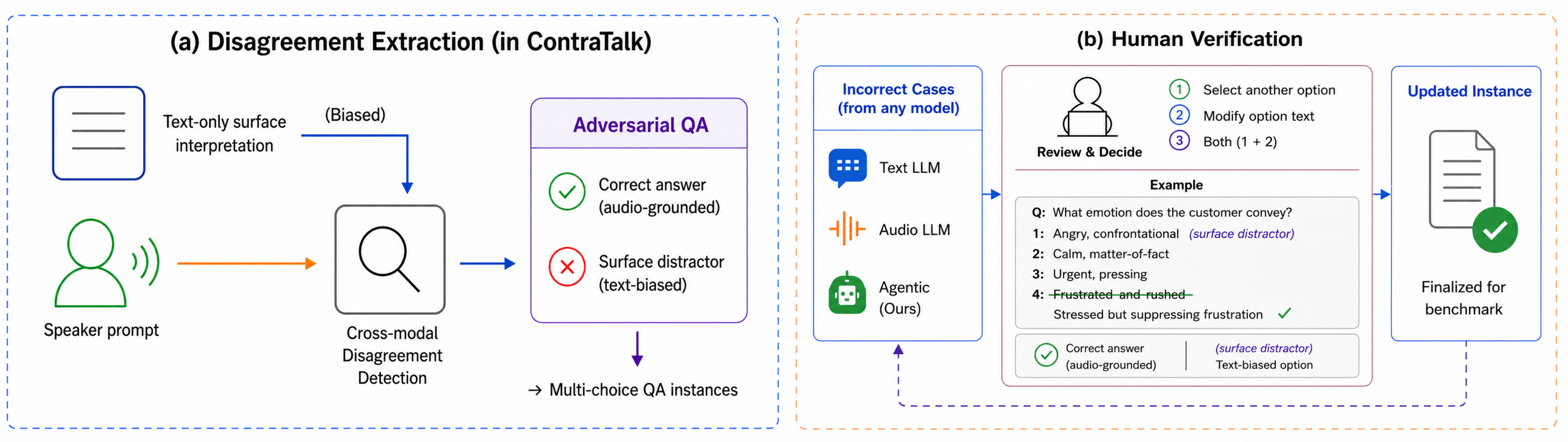}
    \caption{ContraTalk Construction via Cross-Modal Disagreement and Human Verification.}
    \label{fig:conflict_engine}
    \vspace{-5mm}
\end{figure}

\subsection{ContraTalk Benchmark Construction}
\label{sec:benchmark_construction}

\noindent\textbf{Data source and construction-time signals.}
We construct \textsc{ContraTalk} from the Seamless Interaction Dataset~\citep{agrawal2025seamless}, which provides spoken dialogues, transcripts, and speaker prompts describing intended speaker states, roles, or communicative behaviors.
We use these prompts only as construction-time signals to propose candidate audio-grounded interpretations and candidate answer options.
They are not provided to models during evaluation.

\noindent\textbf{Construction overview.}
The goal of the construction process is to create multiple-choice QA instances that evaluate two complementary settings: cases where speech evidence should correct a transcript-based interpretation, and cases where the transcript-supported interpretation should be preserved.
As summarized in Figure~\ref{fig:conflict_engine}, we first derive transcript-based surface interpretations and speaker-conditioned audio-grounded interpretations, then construct examples based on their relationship.
In the \emph{conflict split}, these two interpretations differ: the transcript supports a plausible but incorrect surface interpretation, while acoustic or interactional evidence supports a different grounded answer.
We therefore include the transcript-based interpretation as a text-biased distractor, allowing us to measure whether models are misled by transcript-only reasoning.
In the \emph{consistent split}, the transcript-based and speech-grounded interpretations agree.
Since the surface interpretation matches the grounded answer, no separate misleading option is defined for this split; consistent cases are evaluated by accuracy only.
All instances are verified through automatic checks and targeted human review to ensure that the final answer is supported by the available evidence and that answer choices are distinct.
Full details of the construction pipeline, including conflict/consistent region selection,
question generation, QA-only leakage filtering, and automatic revision are provided in Appendix~\ref{app:benchmark_construction}.

\noindent\textbf{Benchmark composition.}
Because the labeled Seamless Interaction test split contains 117 spoken dialogues, \textsc{ContraTalk} uses all available test dialogues and derives 501 multiple-choice questions from them.
The questions cover five discourse-level dimensions: interaction behavior, emotion state, social stance, dialogue act, and conversational intent.
The benchmark includes both conflict cases, where acoustic or interactional cues are needed to resolve cross-modal disagreement, and consistent cases, where the transcript-supported interpretation should be preserved.
Table~\ref{tab:benchmark_statistics} summarizes the distribution across discourse dimensions.

\noindent\textbf{Human verification.}
We perform targeted human verification through model-disagreement review and sampling-based quality control.
Each candidate instance is evaluated by complementary systems, including a direct Audio-LLM and our agentic-style Audio Twin system; disagreement cases are prioritized for manual review, while agreement cases are sampled for auditing.
In total, 350 of the 501 final examples are manually reviewed by seven reviewers, who listen to the audio and inspect the transcript, question, answer choices, and final answer while remaining blind to the speaker prompts.
Reviewers verify audio grounding, transcript-only distractor plausibility, and answer-option distinctiveness.
Flagged cases are revised, removed, or collected for a unified second-pass review.
The verification protocol and full evaluator instructions are provided in
Appendices~\ref{app:human_verification} and~\ref{app:evaluator_instructions}.


 \begin{table}[t]
\centering
\small
\setlength{\tabcolsep}{4pt}
\begin{tabular}{lllr}
\toprule
Abbr. & Category & Description &  \makebox[0pt][r]{\# Conflict/Consistent} \\
\midrule
IB & Interaction Behavior 
& Turn-taking, hesitation, overlap, or interruption 
& 68/34 \\
ES & Emotion State 
& Affect such as anger, calmness, stress, or frustration 
& 75/38\\
CI & Conversational Intent 
& Implied communicative goal or intention 
& 67/34 \\
DA & Dialogue Act 
& Utterance function, e.g., request or refusal 
& 56/28 \\
SS & Social Stance 
& Social attitude, e.g., politeness or authority
& 67/34 \\
\midrule
Ovr. & Overall 
& Total number of benchmark examples 
& 333/168 \\
\bottomrule
\end{tabular}
\vspace{5pt} 
\caption{
\textbf{Benchmark composition.}
Number of examples in each discourse-level category.
}
\vspace{-5mm} 
\label{tab:benchmark_statistics}
\end{table}


\section{Agentic-style Inference with Audio Twin}

We introduce an agentic-style framework that augments the lexical transcript with a transcript-aligned \emph{Audio Twin}, making speech-derived evidence explicit and available for comparison with transcript-based interpretations before answer selection.

\subsection{Overview}
Building on the QA formulation in Section~\ref{sec:problem_formulation}, our pipeline performs audio-grounded inference in two steps.
It first constructs an Audio Twin, a transcript-aligned evidence structure that links text-readable speech cues to dialogue turns and speakers.
Given a question, it then retrieves relevant Audio Twin evidence and conditions answer prediction on both the transcript and retrieved speech-derived evidence.
%
Given a transcript \(T\), Audio Twin evidence \(E_{\mathrm{AT}}\), question \(q\), and answer candidates \(\mathcal{C}\), the retrieval module returns question-relevant evidence \(E_q \subseteq E_{\mathrm{AT}}\). 
The final answer is predicted as
\[
\hat{y}=f(T,E_q,q,\mathcal{C}).
\]

Here, \(E_q\) is not generic external context, but transcript-aligned acoustic evidence for comparing lexical and speech-grounded interpretations. 
It exposes paralinguistic cues in a text-readable form, allowing the reasoning model to condition on both what was said and how it was said. 
This distinction is central to \textsc{ContraTalk}: the transcript may support a plausible surface answer, while the correct answer requires grounding in acoustic evidence.

\subsection{Audio Twin Representation}
\label{sec:audio_twin_representation}

We instantiate the Audio Twin as a structured, text-readable representation of a spoken dialogue.
Given a dialogue \(D=(T,A)\), where \(T\) is the timestamped transcript and \(A\) is the corresponding audio, speech front ends extract acoustic and paralinguistic features \(F\) and align them to transcript turns.
An Audio Twin builder converts the transcript and aligned speech features into a question-independent evidence structure:
\[
E_{\mathrm{AT}} = \phi_{\mathrm{AT}}(T,F),
\]
where \(E_{\mathrm{AT}}\) denotes the structured evidence used for downstream retrieval and reasoning.

The core goal of Audio Twin is to avoid modality collapse: rather than allowing the reasoning model to default to the transcript or treat audio as an opaque input, Audio Twin exposes speech delivery as explicit evidence that can be compared against transcript-based interpretations.
It represents prosodic, affective, fluency, timing, and overlap cues as compact textual evidence linked to dialogue turns and speakers.
When speaker normalization is meaningful, cues are expressed relative to speaker-specific baselines, allowing the model to reason about whether a speaker sounds louder, faster, or more hesitant than usual.
Implementation details, including evidence-card fields, alignment reliability, and feature textualization, are provided in Appendix~\ref{app:audio_twin_details}.

\begin{figure}[t]
    \centering
    \includegraphics[width=1.0\linewidth]{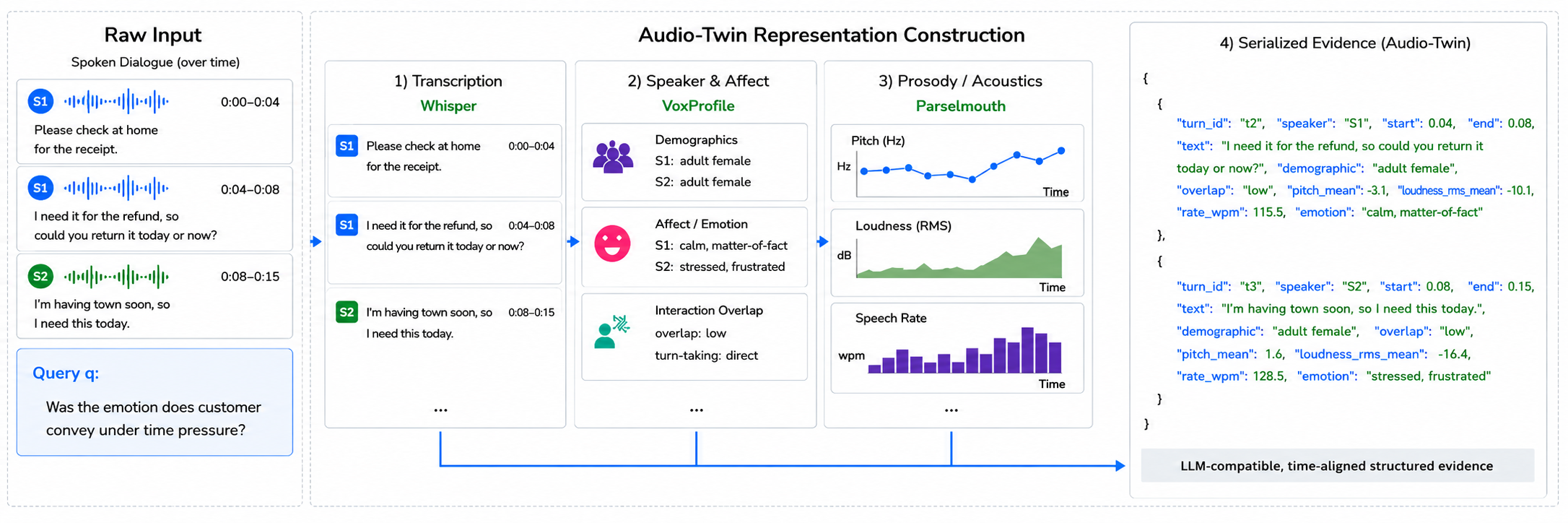}
    \caption{Structured acoustic evidence used by the agentic-style reasoner. Audio Twin converts speech into time-aligned, LLM-compatible evidence entries that expose paralinguistic and interactional cues relevant to the question.}
    \label{fig:structured_evidence}
\vspace{-5mm}
\end{figure}

\subsection{Agentic-style Evidence Retrieval}
\label{sec:agentic_evidence_retrieval}

Given a question \(q\) and the target discourse dimension, the retrieval module constructs an \emph{evidence plan} that specifies what speech-grounded evidence should be retrieved before answer prediction. 
The plan defines the needed evidence types, while a transcript locator identifies the dialogue turns that should serve as anchors.
Inference proceeds in three stages. 
First, the transcript locator selects a small set of relevant transcript anchors without choosing an answer. 
Second, the retrieval module uses these anchors and the evidence plan to fetch the corresponding Audio Twin evidence, including aligned turn-level evidence, speaker baselines, local context, and interaction-level summaries when needed. 
Third, the reasoning model compares transcript-based interpretations with the retrieved speech-grounded evidence and selects the answer most directly supported by the evidence bundle.
For retrieval, let \(U=\{u_i\}_{i=1}^{N}\) denote the set of transcript turns induced by transcript \(T\), and let \(\pi_q\) denote the question-conditioned evidence plan.
The plan selects relevant transcript anchors
\[
S_q = \pi_q(U,q), \qquad S_q \subseteq U.
\]
The retrieval module then returns question-relevant Audio Twin evidence, which is used for answer prediction:
\[
E_q=\mathcal{R}(E_{\mathrm{AT}},S_q,\pi_q), 
\qquad
\hat{y}=f(T,E_q,q,\mathcal{C}).
\]
Here, \(E_q\) contains the Audio Twin entries aligned to the selected anchors, while \(T\) provides the full lexical context.
Details are provided in Appendix~\ref{app:agentic_retrieval_details}.

\section{Experiments}

We organize our evaluation around three research questions:

\noindent\textbf{RQ1: Does \textsc{ContraTalk} expose transcript shortcuts on conflict cases?}
    We test whether transcript-only models are misled when acoustic evidence is needed to resolve cross-modal conflict.

\noindent\textbf{RQ2: Do models preserve performance on consistent cases?}    We test whether audio-capable or agentic-style systems preserve transcript-supported answers when no cross-modal correction is needed.

\noindent\textbf{RQ3: How does audio grounding change model behavior?}
    We compare text-only inference, direct audio inference, and Audio Twin reasoning across both splits to analyze grounding gains and new failure modes.

\subsection{Experimental Setup}
\label{sec:experimental_setup}

\noindent\textbf{Evaluation protocol.}
Each \textsc{ContraTalk} instance consists of dialogue audio $A$, transcript $T$, question $q$, and answer candidates $\mathcal{C}=\{c_1,\ldots,c_K\}$. 

\begin{table*}[t]
\centering
\small
\setlength{\tabcolsep}{3pt}
\begin{tabular}{lrrrrrr|rrrrrr}
\toprule
&  \multicolumn{6}{c|}{Accuracy $\uparrow$} 
& \multicolumn{6}{c}{Mislead Rate $\downarrow$} \\
\cmidrule(lr){2-7} \cmidrule(lr){8-13}
System  
& IB & ES & CI & DA & SS & Ovr.
& IB & ES & CI & DA & SS & Ovr. \\
\midrule
\multicolumn{13}{l}{\emph{Text-only LLMs}} \\
DeepSeek V3.1 
& 42.6 & 34.7 & 37.3 & 42.9 & 31.3 & 37.5
& 33.8 & 41.3 & 44.8 & 33.9 & 38.8 & 38.7 \\
Nova 2 Lite 
& 42.6 & 37.3 & 35.8 & 41.1 & 40.3 & 39.3
& 29.4 & 44.0 & 43.3 & 32.1 & 26.9 & 35.4 \\
Haiku 4.5 
& 36.8 & 26.7 & 28.4 & 44.6 & 31.3 & 33.0
& 47.1 & 53.3 & 50.7 & 37.5 & 34.3 & 45.0 \\
Sonnet 4.5 
& 57.4 & 40.0 & 47.8 & 50.0 & 44.8 & 47.7
& 29.4 & 41.3 & 38.8 & 32.1 & 29.9 & 34.5 \\
Opus 4.7 
& \textbf{61.8} & 40.0 & 40.3 & 48.2 & 37.3 & 45.3
& \textbf{20.6} & 46.7 & 44.8 & 32.1 & 38.8 & 36.9 \\
\midrule
\multicolumn{13}{l}{\emph{Audio-LLMs}} \\
AudioFlamingoNext 
& 45.6 & 50.7 & 41.8 & 50.0 & 43.3 & 46.2
& 29.4 & 37.3 & 37.3 & \textbf{25.0} & 29.9 & 32.1 \\
StepAudio-R1 
& 42.6 & 45.3 & 47.8 & 50.0 & \textbf{49.3} & 46.8
& 36.8 & 30.7 & 35.8 & 30.4 & 26.9 & 32.1 \\
StepAudio-2 
& 44.1 & 38.7 & 37.3 & 35.7 & 40.3 & 39.3
& 22.1 & 34.7 & 32.8 & 39.3 & \textbf{20.9} & 29.7 \\
MIMOAudio 
& 45.6 & 44.0 & 40.3 & 46.4 & 47.8 & 44.7
& 29.4 & 33.3 & 31.3 & 32.1 & 29.9 & 31.2 \\
Qwen2.5-Omni 
& 44.1 & 48.0 & 49.3 & 39.3 & 41.8 & 44.7
& 33.8 & 37.3 & 31.3 & 39.3 & 25.4 & 33.3 \\
KimiAudio 
& 42.6 & \textbf{53.3} & 37.3 & 35.7 & 34.3 & 41.1
& 23.5 & \textbf{29.3} & 40.3 & 28.6 & 29.9 & 30.3 \\
Qwen3-Omni 
& 47.1 & 34.7 & 43.3 & 39.3 & 38.8 & 40.5
& 25.0 & 44.0 & 41.8 & 26.8 & 31.3 & 34.2 \\
GPT-4o-Audio-Mini 
& 35.3 & 36.0 & 28.4 & 32.1 & 35.8 & 33.6
& 36.8 & 40.0 & 46.3 & 42.9 & 34.3 & 39.9 \\

\midrule
\multicolumn{13}{l}{\emph{Audio Twin Reasoning}}\\
Haiku 4.5 + AT
& 39.7 & 37.3 & 46.3 & 51.8 & 43.3 & 43.2
& 22.1 & 38.7 & 35.8 & 26.8 & 29.9 & 30.9 \\
Sonnet 4.5 + AT
& 50.0 & 46.7 & \textbf{55.2} & \textbf{55.4} & 46.3 & \textbf{50.5}
& 27.9 & 37.3 & \textbf{28.4} & 26.8 & 25.4 & \textbf{29.4} \\
Opus 4.7 + AT
& 51.5 & 36.0 & 52.2 & 51.8 & 44.8 & 46.8
& 36.8 & 38.7 & 32.8 & 26.8 & 37.3 & 34.8 \\

\bottomrule
\end{tabular}
\caption{
\textbf{Shortcut diagnostic.}
We report category-wise accuracy and mislead rate across text-only LLMs, direct Audio-LLMs, and Audio Twin reasoning systems.
Mislead rate measures the selection of the transcript-biased surface distractor. 
}
\label{tab:rq1_shortcut_diagnostic}
\vspace{-5mm}
\end{table*}

\noindent\textbf{Models.}
We evaluate \textbf{text-only LLMs}, \textbf{Audio-LLMs}, and \textbf{Audio Twin systems}, covering transcript-only inference, direct speech inference, and structured acoustic-evidence reasoning.
When possible, we compare matched text-only and audio-capable variants to isolate the effect of modality access.
Full model details are provided in Appendix~\ref{app:model_details}.

\noindent\textbf{Metrics.}
We report accuracy and mislead rate.
Accuracy measures whether the model selects the gold answer.
Mislead rate measures selection of the text-biased surface distractor and is reported for conflict cases, where such a distractor is defined.
For consistent cases, we report accuracy.
Metrics are reported overall and by discourse-level category.
We report dialogue-level bootstrap 95\% confidence intervals in
Appendix~\ref{app:statistical-uncertainty}.

\subsection{Experimental Results}

\noindent\textbf{RQ1: Does \textsc{ContraTalk} expose transcript shortcuts on conflict cases?}
We first evaluate whether the conflict split of \textsc{ContraTalk} exposes transcript-based shortcut behavior.
Table~\ref{tab:rq1_shortcut_diagnostic} reports category-wise accuracy and mislead rate on cases where the transcript supports a plausible surface interpretation, while speech-grounded evidence supports a different answer.
Text-only LLMs perform modestly on conflict cases but frequently select the transcript-biased distractor.
Overall, they obtain 33.0--47.7\% accuracy while choosing the mislead option in 34.5--45.0\% of examples.
The shortcut is especially visible for emotion state and conversational intent, where several text-only models show mislead rates above 40\%.
This indicates that the conflict split exposes a systematic shortcut: the transcript alone supports a plausible but incorrect interpretation that can compete with the speech-grounded answer.
Direct Audio-LLMs only partially reduce this failure mode.
Their accuracy ranges from 33.6\% to 46.8\%, while mislead rates remain high at 29.7--39.9\%, indicating that raw audio access alone does not guarantee grounding when transcript and speech cues disagree.
Audio Twin reasoning provides the strongest overall shortcut resistance among the evaluated systems.
Across LLM backbones, Audio Twin variants reach 43.2--50.5\% accuracy with 29.4--34.8\% mislead rate.
Sonnet 4.5 with Audio Twin achieves the best overall accuracy and lowest overall mislead rate, with strong category-level performance on conversational intent and dialogue act.
Nevertheless, the remaining misled cases show that Audio Twin reasoning mitigates, but does not eliminate, transcript-induced shortcuts.
Overall, the conflict split exposes a targeted cross-modal failure mode: models may prefer transcript-plausible answers even when they conflict with how the dialogue is spoken.

\begin{table*}[t]
\centering
\small
\setlength{\tabcolsep}{3pt}
\begin{tabular}{p{2.80cm}cccccc|p{0.84cm}cccccc}
\toprule
& \multicolumn{6}{c|}{Accuracy $\uparrow$}
& & \multicolumn{6}{c}{Accuracy $\uparrow$} \\
\cmidrule(lr){2-7} \cmidrule(lr){9-14}
System & IB & ES & CI & DA & SS & \makebox[0pt][c]{Ovr.}
&  \makebox[0pt][l]{System} & IB & ES & CI & DA & SS & \makebox[0pt][c]{Ovr.} \\
\midrule
\multicolumn{7}{l|}{\emph{Text-only LLMs}} & \multicolumn{7}{l}{\emph{Text-only LLMs}} \\
DeepSeek V3.1 
& 82.4 & 94.7 & 94.1 & 92.9 & 94.1 & 91.7
& Haiku & 88.2 & 94.7 & 97.1 & 96.4 & 88.2 & 92.9 \\
Nova 2 Lite 
& 85.3 & 92.1 & 94.1 & 89.3 & {97.1} & 91.7
& Sonnet & 91.2 & {97.4} & {100} & {100} & 94.1 & 96.4 \\
& & & & & &
& Opus
& {97.1} & {97.4} & {100} & {100} & {97.1} & \textbf{98.2} \\

\midrule
\multicolumn{7}{l|}{\emph{Audio-LLMs}}
& \multicolumn{7}{l}{\emph{Audio Twin Reasoning}} \\
AudioFlamingoNext 
& 85.3 & 89.5 & 91.2 & 71.4 & {97.1} & 87.5 
& Haiku & 85.3 & 89.5 & 64.7 & 89.3 & 79.4 & 81.5\\
StepAudio-R1 
& 82.4 & 92.1 & 91.2 & 89.3 & 94.1 & \textbf{89.9}
& Sonnet
& {97.1} & 78.9 & 88.2 & 96.4 & 85.3 & 88.7 \\
StepAudio-2 
& 61.8 & 65.8 & 73.5 & 60.7 & 85.3 & 69.6
& Opus & 94.1 & 94.7 & 94.1 & 96.4 & 91.2 & \textbf{94.0} \\

MIMOAudio 
& 82.4 & 84.2 & 88.2 & 67.9 & 88.2 & 82.7
&  &  &  &  &  &  \\
Qwen2.5-Omni 
& 79.4 & 89.5 & 76.5 & 85.7 & 88.2 & 83.9
& 
&  &  &  &  &  &  \\
KimiAudio 
& 64.7 & 86.8 & 88.2 & 67.9 & 82.4 & 78.6
&  
&  &  &  &  &  &  \\
Qwen3-Omni 
& 85.3 & 81.6 & 76.5 & 78.6 & 94.1 & 83.3
&  
&  &  &  &  &  &  \\
\makebox[0pt][l]{GPT-4o-Audio-Mini} 
& 91.2 & 76.3 & 76.5 & 75.0 & {97.1} & 83.3
&  
&  &  &  &  &  &  \\
\bottomrule
\end{tabular}
\caption{
\textbf{Consistent-split performance.}
We report category-wise accuracy on examples where transcript-supported and audio-grounded answers are aligned.
No mislead rate is reported because no transcript-biased distractor is defined for this split.
}
\label{tab:rq2_consistent_split_accuracy}
\vspace{-5mm}
\end{table*}

\noindent\textbf{RQ2: Do models preserve performance on consistent cases?}
We next evaluate whether models remain reliable on consistent cases, where transcript-supported and speech-grounded interpretations agree.
Table~\ref{tab:rq2_consistent_split_accuracy} reports category-wise accuracy. 
Because no text-biased surface distractor is defined, we report accuracy only.
Models perform substantially better on consistent cases than on conflict cases.
Strong text-only LLMs often exceed 90\% accuracy, with Opus 4.7 reaching 98.2\% overall.
Their performance is especially high on conversational intent and dialogue act, where the strongest text-only models reach 100\% accuracy.
This suggests that \textsc{ContraTalk} is not simply difficult due to ambiguous questions or noisy choices; the main challenge arises when transcript and speech cues diverge.
Direct Audio-LLMs achieve high accuracy on many consistent examples, but their performance is less uniform across systems.
For example, StepAudio-R1 reaches 89.9\% overall, while StepAudio-2 drops to 69.6\%.
The drop is particularly visible in interaction behavior and dialogue acts for weaker Audio-LLMs, suggesting that speech access does not always preserve transcript-supported interpretations.
Audio Twin reasoning shows a similar backbone-dependent pattern: Opus 4.7 + AT maintains strong, consistent-case accuracy at 94.0\%, whereas smaller backbones degrade more.
Together with the conflict-split results, this comparison shows that \textsc{ContraTalk} separates two behaviors: preserving correct interpretations when modalities agree, and revising transcript-based interpretations when they conflict.


\begin{table*}[t]
\centering
\small
\setlength{\tabcolsep}{4pt}
\begin{tabular}{lccc|ccc}
\toprule
& \multicolumn{3}{c|}{Text-only} 
& \multicolumn{3}{c}{Speech-grounded} \\
\cmidrule(lr){2-4} \cmidrule(lr){5-7}
System 
& Conf. Acc. $\uparrow$ & Mis. $\downarrow$ & Cons. Acc. $\uparrow$
& Conf. Acc. $\uparrow$ & Mis. $\downarrow$ & Cons. Acc. $\uparrow$ \\
\midrule
\multicolumn{7}{l}{\emph{Direct Audio-LLMs: Text vs. Audio}} \\
StepAudio-R1 
& 42.3 & 34.5 & 91.7 & 46.8 & 32.1 & 89.9 \\
StepAudio-2 
& 34.2 & 37.8 & 83.9 & 39.3 & 29.7 & 69.6 \\
Qwen3-Omni 
& 42.9 & 40.5 & 91.1 & 40.5 & 34.2 & 83.3 \\
KimiAudio 
& 43.2 & \textbf{28.8} & 84.5 & 41.1 & 30.3 & 78.6 \\
MIMOAudio 
& 41.7 & 33.6 & 78.0 & 44.7 & 31.2 & 82.7 \\
Qwen2.5-Omni 
& 41.1 & 32.7 & 92.9 & 44.7 & 33.3 & 83.9 \\

\midrule
\multicolumn{7}{l}{\emph{Agentic-style LLMs: Text vs. Audio Twin}} \\
Haiku 4.5 
& 33.0 & 45.0 & 92.9 & 43.2 & 30.9 & 81.5 \\
Sonnet 4.5 
& \textbf{47.7} & 34.5 & 96.4 & \textbf{50.5} & \textbf{29.4} & 88.7 \\
Opus 4.7 
& 45.3 & 36.9 & \textbf{98.2} & 46.8 & 34.8 & \textbf{94.0} \\

\bottomrule
\end{tabular}
\caption{
\textbf{Paired text-only and speech-grounded inference.}
Results from Tables~\ref{tab:rq1_shortcut_diagnostic} and~\ref{tab:rq2_consistent_split_accuracy} are reorganized into matched text-only and speech-grounded pairs, enabling direct comparison across conflict and consistent splits.
}\label{tab:rq3_paired_grounding_comparison}
\vspace{-5mm}
\end{table*}

\noindent\textbf{RQ3: How does audio grounding change model behavior?}
Table~\ref{tab:rq3_paired_grounding_comparison} combines the conflict results from Table~\ref{tab:rq1_shortcut_diagnostic} and the consistent-split results from Table~\ref{tab:rq2_consistent_split_accuracy}, but reorganizes them as matched text-only and speech-grounded pairs.
This paired view allows us to examine whether adding speech helps resolve transcript-based shortcuts on conflict cases without disrupting transcript-supported reasoning on consistent cases.
The paired comparison reveals a modality-collapse pattern.
Several direct Audio-LLMs improve conflict-case behavior, either increasing conflict accuracy or reducing mislead rate, but their accuracy drops on the consistent split after speech is introduced.
This suggests that speech input can shift model predictions away from transcript-biased shortcuts, but may also destabilize cases where the transcript already supports the correct answer.
Other Audio-LLMs show limited improvement on conflict cases and similar mislead rates, suggesting continued reliance on transcript-based cues despite audio input.
Thus, direct audio access does not guarantee calibrated multimodal reasoning, either improving conflict cases at the cost of consistency or showing limited corrective use of audio.
Our Audio Twin framework improves conflict-case accuracy while reducing trap selection, but its consistent-case behavior remains backbone-dependent: smaller backbones are more easily perturbed by added acoustic evidence, while stronger backbones better preserve transcript-supported predictions.
Overall, effective speech grounding should correct transcript-based shortcuts under conflict while preserving transcript-supported reasoning when modalities agree.

\section{Conclusion}

We introduced \textsc{ContraTalk}, a benchmark for evaluating spoken dialogue understanding under cross-modal disagreement. Unlike evaluations that can be solved from transcripts alone or from audio alone, \textsc{ContraTalk} separates conflict cases, where transcript-based interpretations are plausible but contradicted by acoustic evidence, from consistent cases, where transcript and speech support the same answer. This design enables us to test not only whether models can use audio to correct transcript shortcuts, but also whether they remain calibrated when no correction is needed.

Our experiments show that strong text-only LLMs perform well on consistent cases but degrade sharply on conflict cases, confirming that transcript-based performance can mask failures in speech-grounded understanding. Direct AudioLLMs provide only partial grounding: speech input sometimes improves conflict-case accuracy, but mislead rates remain substantial and consistent-case accuracy can drop when audio is introduced. Our agentic-style Audio Twin framework provides a more explicit interface to acoustic evidence and improves conflict-case behavior for several backbones. At the same time, the consistent-case results show that explicit audio grounding must be applied selectively rather than uniformly.

These findings suggest that robust spoken dialogue understanding requires more than implicit multimodal fusion, while also highlighting that explicit acoustic evidence must be used selectively. Models must learn to decide when transcript evidence is reliable, when acoustic cues should revise a surface reading, and how to incorporate speech-derived evidence without disrupting otherwise correct text-supported reasoning. We hope \textsc{ContraTalk} supports future work on disagreement-aware evaluation, calibrated acoustic grounding, and reasoning architectures that make speech evidence more explicit, auditable, and reliable.

\paragraph{Limitations}
This work focuses on controlled cross-modal disagreement rather than the full range of cues in real-world spoken dialogue understanding. Audio Twin is one instantiation of explicit audio-grounded reasoning; further discussion is provided in Appendix~\ref{app:limitations}.

\begin{ack}
This work was supported in part by the Johns Hopkins University + Amazon Initiative for Artificial Intelligence (AI2AI) Fellowship. We also thank Michael Owen for valuable discussions and insightful feedback.


\end{ack}

\appendix

\section{Limitations}
\label{app:limitations}
This work focuses on evaluating and mitigating transcript shortcuts under controlled cross-modal disagreement, rather than providing a complete solution to spoken dialogue understanding. Although \textsc{ContraTalk} covers multiple discourse-level dimensions, it does not exhaust the full range of acoustic, pragmatic, social, and cultural cues that may shape real-world conversations.

Our Audio Twin framework is also only one instantiation of explicit audio-grounded reasoning. More complex agentic systems, stronger speech encoders, or improved evidence selection strategies may further improve the ability to reconcile conflicting modalities.

Finally, Audio Twin provides an interpretable interface to acoustic evidence, but its utility depends on the fidelity of the extracted speech cues. Automatic transcription, alignment, prosodic analysis, and affective cue estimation may introduce noise, which can limit reasoning performance when the relevant evidence is subtle, ambiguous, or difficult to localize.

\section{ContraTalk Benchmark Construction Details}
\label{app:benchmark_construction}

\begin{figure*}[h]
    \centering
    \includegraphics[width=.85\textwidth]{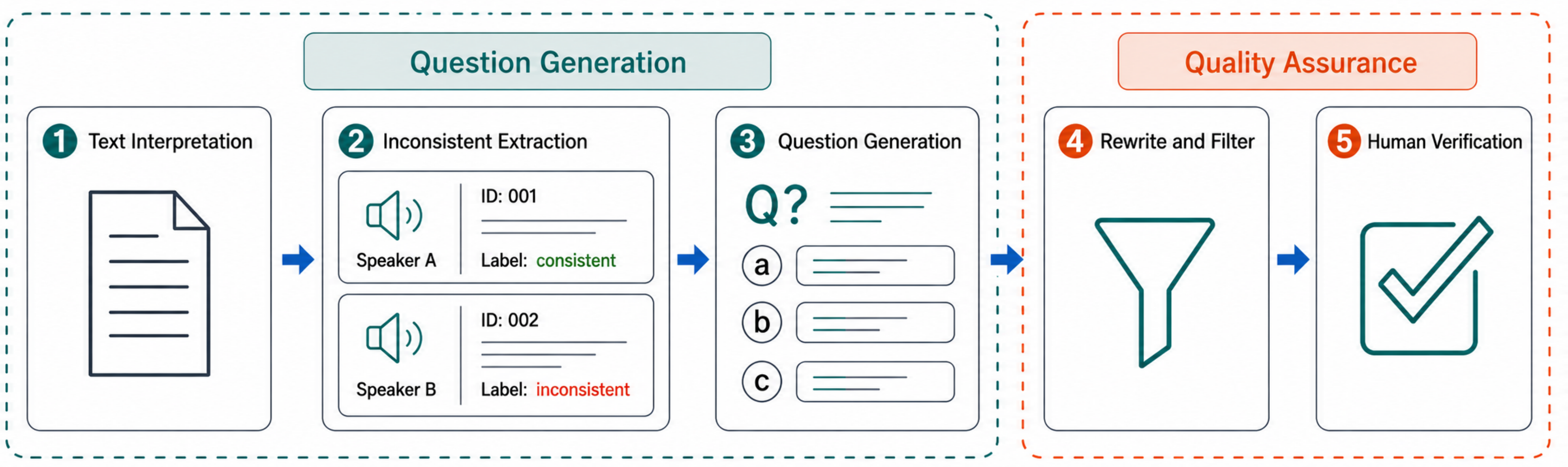}
    \caption{Overview of the \textsc{ContraTalk} benchmark construction pipeline. We first capture text-only surface interpretations, then identify disagreement regions, generate question-answer pairs, and perform multiple stages of quality assurance.}
    \label{fig:pipeline_overview}
\end{figure*}

This appendix provides the full construction and verification procedure for \textsc{ContraTalk}. 
The main paper summarizes the overall pipeline in Section~\ref{sec:benchmark_construction}; here we describe each step in detail. The overview is shown in Fig.~\ref{fig:pipeline_overview}

\paragraph{Step 1: Text-only surface interpretation.}
We first estimate the interpretation that a text-only reasoner would form from the transcript alone. 
Given a dialogue transcript, we prompt a language model to annotate utterances along five discourse-level dimensions: \emph{interaction behavior, emotion state, social stance,  dialogue act}, and \emph{conversational intent}. 
This step captures transcript-induced surface bias and serves as the comparison point for identifying text--audio disagreement.

\paragraph{Step 2: Speaker-conditioned label comparison.}
We compare the text-only surface interpretations with speaker-conditioned candidate grounded labels derived from the dataset prompts.
The prompts are used only as construction-time signals to propose candidate audio-grounded labels; they are not provided to models during evaluation, and final labels are assigned after human verification.
This comparison produces two types of candidate regions.
First, we identify \emph{conflict} regions where the transcript-based interpretation is inconsistent with, underspecified relative to, or misleading compared with the intended spoken realization.
Second, we identify \emph{consistent} regions where the transcript-based interpretation is closely aligned with the speaker-conditioned grounded label.
The former supports construction of audio-dependent conflict questions, while the latter supports construction of text-solvable consistent questions.

\paragraph{Step 3: Conflict and consistent question generation.}
We convert the selected regions into multiple-choice QA instances under two regimes.
For conflict cases, each question targets a discourse-level dimension and asks for the grounded interpretation of an utterance or dialogue segment.
The correct answer reflects the intended spoken realization, while a text-biased distractor is designed to remain plausible under transcript-only reasoning.
This makes the incorrect option a structured trap rather than a random negative.
For consistent cases, we generate questions from regions where transcript-based and speaker-conditioned interpretations agree.
The correct answer is supported by both the transcript and the spoken realization, and the distractors are constructed to be clearly distinguishable without relying on cross-modal disagreement.
These consistent questions test whether models preserve transcript-supported interpretations when no audio-based correction is required.

\paragraph{Step 4: Automatic quality control.}
Automatically generated questions may contain artifacts that reveal the answer without requiring dialogue evidence.
We therefore first run each instance in a QA-only setting, where the model receives only the question and answer candidates, without transcript or audio, and is asked to provide both an answer and its reasoning.
We audit the QA-only reasoning to determine whether the answer can be justified solely from the question and answer candidates.
If the reasoning reveals a valid dialogue-independent shortcut, we rewrite the question and rerun the QA-only test in a second round.
Instances that remain answerable without dialogue evidence after revision are removed.

\paragraph{Step 5: Human verification.}
Finally, we manually verify candidate instances for answerability, evidence quality, and distractor validity.
For conflict cases, we prioritize review using two automatic validators: a direct Audio-LLM and an Audio Twin pipeline.
Validator-disagreement cases are treated as higher-risk cases and manually inspected.
For consistent cases, where no text-biased surface distractor is defined, we instead conduct random sampling-based quality control.
Annotators listen to the audio and revise or remove instances with unclear evidence, ambiguous answers, or indistinguishable answer choices.

\section{Statistical Uncertainty}
\label{app:statistical-uncertainty}

Table~\ref{tab:cluster-bootstrap-ci} reports aggregate confidence intervals for the main metrics used in Tables~2--4.

We estimate uncertainty using a nonparametric dialogue-level bootstrap. 
For each system and evaluation split, we sample dialogues with replacement, include all questions derived from each sampled dialogue, and recompute accuracy and, for conflict cases, mislead rate. 
We report 95\% percentile confidence intervals from 10{,}000 bootstrap samples, using the 2.5th and 97.5th percentiles of the resulting metric distribution. The reported intervals are percentile confidence intervals, not standard deviations or
standard errors, and the procedure does not assume a normal error distribution.

Because multiple questions may come from the same dialogue, we resample dialogues rather than individual questions. 
These intervals quantify score variability under dialogue-level resampling of \textsc{ContraTalk}; they do not account for uncertainty from alternative benchmark construction or labeling decisions.

\begin{table}[t]
\centering
\footnotesize
\setlength{\tabcolsep}{4pt}
\begin{tabular}{@{}lccc@{}}
\toprule
\textbf{System} &
\textbf{Conflict Acc.} &
\textbf{Conflict Mislead} &
\textbf{Consistent Acc.} \\
\midrule
\multicolumn{4}{@{}l}{\textit{Text-only LLMs}} \\
DeepSeek V3.1       & 37.5 [32.4, 42.7] & 38.7 [33.9, 43.5] & 91.7 [87.1, 95.7] \\
Nova 2 Lite         & 39.3 [33.5, 45.4] & 35.4 [30.1, 40.7] & 91.7 [87.3, 95.5] \\
Haiku 4.5           & 33.0 [27.8, 38.3] & 45.0 [39.1, 50.9] & 92.9 [89.0, 96.2] \\
Sonnet 4.5          & 47.7 [41.0, 54.1] & 34.5 [28.1, 41.3] & 96.4 [93.4, 98.8] \\
Opus 4.7            & 45.3 [39.4, 51.3] & 36.9 [31.1, 43.0] & 98.2 [96.0, 100.0] \\
KimiAudio           & 43.2 [38.2, 48.2] & 28.8 [24.4, 33.3] & 84.5 [79.3, 89.6] \\
MIMOAudio           & 41.7 [36.1, 47.4] & 33.6 [28.6, 39.2] & 78.0 [71.9, 83.7] \\
Qwen2.5-Omni        & 41.1 [35.9, 46.4] & 32.7 [27.8, 37.8] & 92.9 [89.0, 96.3] \\
Qwen3-Omni          & 42.9 [36.8, 49.2] & 40.5 [34.4, 46.7] & 91.1 [87.0, 94.8] \\
StepAudio-2         & 34.2 [28.9, 39.5] & 37.8 [32.2, 43.6] & 83.9 [77.8, 89.7] \\
StepAudio-R1        & 42.3 [36.6, 48.1] & 34.5 [28.8, 40.4] & 91.7 [87.5, 95.7] \\
\midrule
\multicolumn{4}{@{}l}{\textit{Audio-LLMs}} \\
AudioFlamingoNext   & 46.2 [40.3, 52.2] & 32.1 [27.5, 37.0] & 87.5 [82.3, 92.3] \\
GPT-4o-Audio-Mini   & 33.6 [28.6, 38.9] & 39.9 [34.1, 46.0] & 83.3 [77.1, 89.0] \\
KimiAudio           & 41.1 [35.1, 47.4] & 30.3 [25.0, 35.8] & 78.6 [72.5, 84.3] \\
MIMOAudio           & 44.7 [39.1, 50.1] & 31.2 [26.2, 36.4] & 82.7 [77.0, 88.0] \\
Qwen2.5-Omni        & 44.7 [38.7, 50.8] & 33.3 [28.0, 39.0] & 83.9 [78.7, 89.0] \\
Qwen3-Omni          & 40.5 [35.2, 46.1] & 34.2 [28.4, 40.2] & 83.3 [76.9, 89.4] \\
StepAudio-2         & 39.3 [33.5, 45.2] & 29.7 [24.6, 35.1] & 69.6 [61.2, 77.3] \\
StepAudio-R1        & 46.8 [40.6, 53.0] & 32.1 [26.6, 37.9] & 89.9 [84.0, 94.8] \\
\midrule
\multicolumn{4}{@{}l}{\textit{Audio Twin Reasoning}} \\
Haiku 4.5 + AT      & 43.2 [37.8, 48.5] & 30.9 [25.6, 36.6] & 81.5 [75.4, 87.2] \\
Sonnet 4.5 + AT     & 50.5 [44.9, 55.9] & 29.4 [25.0, 34.1] & 88.7 [83.6, 93.5] \\
Opus 4.7 + AT       & 46.8 [41.0, 52.7] & 34.8 [29.6, 39.9] & 94.0 [89.3, 97.7] \\
\bottomrule
\end{tabular}
\vspace{10pt}
\caption{
Dialogue-level bootstrap 95\% confidence intervals for aggregate model
performance. Entries are point estimates with 95\% percentile confidence
intervals in brackets, reported as percentages. The conflict split contains
333 questions from 113 dialogues; the consistent split contains 168 questions
from 75 dialogues. Split-specific dialogue counts do not sum to the total number
of dialogues because a dialogue may contribute questions to both splits. Mislead
rate is defined only for conflict cases. AT denotes Audio Twin.
}
\label{tab:cluster-bootstrap-ci}
\end{table}

\section{Human Verification Protocol}
\label{app:human_verification}

We use a verification protocol that combines targeted review for conflict cases with sampling-based quality control for consistent cases.
Before human review, each conflict candidate is evaluated by complementary systems, including a text-only LLM, a direct Audio-LLM, and our agentic-style Audio Twin reasoning system.
Instances for which the systems disagree are treated as higher-risk cases and manually reviewed.
For consistent candidates, where the transcript and speech evidence support the same answer and no text-biased surface distractor is defined, we conduct random sampling-based review to check overall quality.
Together, this procedure results in 350 manually reviewed instances out of the final 501 examples.
Table~\ref{tab:human-verification-summary} summarizes the verification coverage and major failure modes.

Human verification is conducted by seven reviewers.
Reviewers are given the dialogue audio, transcript, question, answer choices, and the proposed final answer, but are blind to the construction-time speaker prompts used during benchmark construction.
They are instructed to listen to the audio and verify whether the selected answer is supported by the dialogue evidence, whether the answer choices are sufficiently distinct, and whether the question is answerable without relying on construction-time metadata.
For conflict cases, reviewers additionally check whether the surface distractor is plausible from the transcript alone while remaining distinguishable from the speech-grounded answer.

Instances are revised or removed if the audio evidence is unclear, the correct answer is not sufficiently grounded, the options are ambiguous, or the distractor is too close to the verified answer.
When reviewers are uncertain, the instance is collected for a second-pass verification round, where the evidence, answer options, and final label are re-examined under a unified guideline.
Only examples that pass this process are retained in the final benchmark; otherwise, they are revised or discarded.

\begin{table}[t]
\centering
\small
\setlength{\tabcolsep}{4pt}
\begin{tabularx}{\linewidth}{lrrX}
\toprule
\textbf{Subset} & 
\textbf{\# Cand.} & 
\textbf{\# Reviewed} & 
\textbf{Major Failure Modes} \\
\midrule
Conflict cases 
& 333 
& 230 
& Ambiguous acoustic evidence; weak surface distractor; unclear audio grounding \\

Consistent cases 
& 168 
& 120 
& Transcript-answer mismatch; answer leakage; near-duplicate options \\

\midrule
Final retained benchmark 
& 501 
& 350 
& -- \\
\bottomrule
\end{tabularx}
\vspace{10pt}
\caption{
Human verification summary for \textsc{ContraTalk}. 
We manually review 230 conflict cases and 120 consistent cases, covering 350 of the 501 retained benchmark examples.
Conflict cases are checked for answerability, speech-grounded evidence, and surface-distractor quality, while consistent cases are checked to ensure that transcript-supported and speech-grounded interpretations are aligned.
}
\label{tab:human-verification-summary}
\end{table}

\section{Instructions Given to Human Evaluators}
\label{app:evaluator_instructions}

This section provides the full instructions shown to human evaluators during verification.

\paragraph{Task description.}
Each instance contains a question, four answer choices, a labeled answer index, a short audio segment of approximately 15 seconds, and a transcript for reference.
Evaluators are asked to verify whether the labeled answer is correct, using the audio as the primary source of evidence.
The transcript may be used as support, but when the audio and transcript suggest different interpretations, evaluators are instructed to trust the audio.

\paragraph{Verification task.}
For each instance, evaluators first listen to the audio and decide whether the labeled answer is correct.
If the labeled answer is correct, they mark the instance as correct.
If the labeled answer is incorrect, they either update the final answer index or revise one of the answer choices.

When the correct answer is already present among the four choices, evaluators simply update the final answer index.
If none of the four choices is fully correct, evaluators revise one choice and provide both the final answer index and the updated choice text.
They are instructed to keep answer choices clear, specific, and mutually distinguishable, so that each option expresses a distinct emotion, intent, stance, or behavior.

\paragraph{Mislead options.}
For conflict cases, each question includes a mislead option, which is an intentionally designed text-biased distractor.
Evaluators are told not to avoid this option simply because it is labeled as a distractor.
If the mislead option is actually supported by the audio, they should select it as the final answer by updating the answer index.

\paragraph{Quality-control guidelines.}
Evaluators are instructed to prioritize audio evidence over transcript-based interpretations, avoid making two answer choices nearly identical, and ensure that the final options remain clearly distinguishable.
Instances are revised when the answer can be corrected by changing the index or editing a choice.
Instances are removed when the audio evidence is unclear, the question is not answerable, or the options cannot be made sufficiently distinct.

\paragraph{Evaluator workflow.}
The verification workflow is as follows:
\begin{enumerate}
    \item Listen to the audio and consult the transcript only if needed.
    \item Decide whether the labeled answer is correct.
    \item If correct, mark the instance as correct.
    \item If incorrect, either update the final answer index or revise one answer choice.
    \item Check that all answer choices are clear and distinguishable.
\end{enumerate}

\section{Audio Twin Representation Details}
\label{app:audio_twin_details}

This appendix provides implementation details for the Audio Twin representation used in Section~\ref{sec:audio_twin_representation}. 
The Audio Twin stores speech-derived evidence as compact JSON-like \emph{evidence cards} with stable identifiers and fixed fields. 
These cards are constructed independently of the downstream question and are queried only during inference.

\noindent\textbf{Feature extraction front ends.}
We use Whisper~\citep{radford2023robust} to obtain timestamped transcripts and turn-level ASR segments. 
Low-level prosodic features, including loudness, fundamental frequency, and pitch variation, are extracted with Parselmouth~\citep{jadoul2018introducing}, a Python interface to Praat. 
Higher-level speaker and speech-trait attributes, including speaker demographics, fluency, valence, arousal, dominance, and categorical emotion, are estimated with Vox-Profile~\citep{feng2025vox}. 
These front ends provide the raw cues that are aligned to transcript turns and textualized into Audio Twin evidence cards.

\paragraph{Turn alignment.}
Each transcript turn is linked to its corresponding acoustic segment using timestamp overlap when available. 
Timestamp alignment is assigned high reliability when the overlap covers at least half of the transcript-turn duration, and medium reliability when positive overlap exists but the ratio is below this threshold. 
When timestamp alignment is unavailable or insufficient, the system falls back to transcript-text similarity with a minimum similarity score of \(0.50\). 
If this also fails, the system uses same-speaker turn order and marks the evidence as low reliability. 
If no same-speaker segment is available, the alignment is marked missing.

\paragraph{Evidence families.}
The Audio Twin contains three main evidence families.
\begin{itemize}
    \item \textbf{Turn cards.} 
    Each turn card links one transcript turn to aligned speech-derived cues, including loudness, pitch level, pitch variation, speaking rate, pause-before, pause-after, overlap, valence, arousal, dominance, categorical emotion, emotion confidence, and fluency. 
    It also stores the alignment method and reliability.

    \item \textbf{Speaker baseline cards.} 
    These cards summarize each speaker's usual loudness, pitch level, pitch variation, speaking rate, affective profile, and fluency. 
    They support speaker-relative interpretation, for example distinguishing generally expressive speakers from speakers who are unusually expressive in a specific turn. 
    Baseline reliability is marked high when at least four same-speaker feature observations are available and low otherwise.

    \item \textbf{Dialogue-dynamics cards.} 
    These cards summarize interaction-level patterns such as turn counts, speaking time, overlap count, total and maximum overlap duration, and mean response-delay patterns for each speaker.
\end{itemize}

The implementation also supports auxiliary evidence records when required by the retrieval plan. 
Local window records preserve nearby transcript turns around a target line. Context blocks attach selected-line functions to the local window and include notes reminding the model to interpret the target line in chronological context. 
Representative-turn cards retrieve high-arousal or high-dominance turns for a speaker. 
Speaker-comparison cards place both speakers' baseline summaries in the same record, allowing the model to compare their typical delivery patterns directly. 
Speaker-attribute records, when present in the feature files, are stored separately from turn-level acoustic evidence.

\paragraph{Evidence identifiers.}
Each evidence entry has a stable identifier. 
Transcript anchors use line identifiers such as \texttt{L015}. 
Turn cards use identifiers such as \texttt{CARD\_L015}; local windows and context blocks use \texttt{WINDOW\_L015} and \texttt{CONTEXT\_L015}. 
Speaker baselines use \texttt{BASELINE\_A} and \texttt{BASELINE\_B}, and dialogue-level evidence uses identifiers such as \texttt{DIALOGUE\_DYNAMICS} and \texttt{COMPARISON\_OVERALL}. 

\paragraph{Feature textualization.}

Continuous acoustic and timing features are converted into text before being exposed to the reasoning model. 
Prosodic and timing values are encoded relative to the same speaker's distribution. 
When at least four same-speaker observations are available, the system uses a \(z\)-score threshold of \(\pm 0.75\) to label a value as lower than usual, typical, or higher than usual. 
When fewer observations are available, it falls back to a ratio comparison against the same-speaker mean: values below \(0.75\times\) the speaker mean are labeled lower than usual, values above \(1.25\times\) the speaker mean are labeled higher than usual, and values in between are labeled typical.

Affective scores are discretized with fixed thresholds on normalized model-output scales. 
For valence scores in the \([0,1]\) range, the system maps values to \emph{very negative}, \emph{negative}, \emph{neutral}, \emph{positive}, and \emph{very positive} using thresholds \(0.30\), \(0.45\), \(0.55\), and \(0.70\). 
For arousal, dominance, and emotion confidence scores in the \([0,1]\) range, the system uses thresholds \(0.40\) and \(0.70\) to produce low, medium, and high labels. 
Categorical emotion predictions are combined with valence, arousal, and dominance to form a compact emotion-impression field.

\paragraph{Evidence notes.}
Turn cards additionally include short, rule-derived notes describing what the acoustic evidence supports, what interpretations it remains compatible with, and what it cannot establish without discourse context. 
For example, high arousal supports high vocal or emotional activation, while negative valence supports negative affect. 
The notes also state limitations: sincerity, social intent, sarcasm, hostility, and confidence require dialogue context and cannot be inferred from low-level acoustic cues alone. 
These notes are intended to make the evidence usable for reasoning while avoiding overclaiming from acoustic features.

\paragraph{Information isolation.}
The Audio Twin is built from the transcript and audio-derived features only. 
It does not include gold answers, speaker prompts, construction-time labels, human-verification decisions, or question-generation explanations. 
The model-visible question object contains only the question text, four answer candidates, and the question category. 
Thus, the representation provides a textual interface to speech delivery rather than a shortcut to the benchmark labels.

\section{Agentic-style Evidence Retrieval Details}
\label{app:agentic_retrieval_details}

This appendix provides implementation details for the agentic-style evidence retrieval procedure described in Section~\ref{sec:agentic_evidence_retrieval}. 
Given a question \(q\), answer candidates \(C=\{c_1,\ldots,c_4\}\), and the target discourse dimension, the system first constructs an evidence plan that determines what evidence should be retrieved before answer prediction.

\paragraph{Evidence-plan assignment.}
For the evaluated target dimensions, evidence-plan assignment is deterministic. 
Questions targeting emotion state are assigned to \emph{emotion state}; questions targeting interaction behavior are assigned to \emph{interaction behavior}; social-stance and conversational-intent questions are assigned to \emph{speaker-level style}; and dialogue-act questions are assigned to \emph{local prosodic delivery}. 
Questions or answer candidates that require comparing speakers, including demographic-choice cues, are assigned to \emph{speaker comparison}. 
The model may propose additional evidence to inspect, but it cannot override the rule-based evidence-plan type. 

\paragraph{Evidence-plan types.}
We use five evidence-plan types, each corresponding to a different kind of speech evidence needed for answering the question.
\begin{itemize}
    \item \textbf{Emotion state.} 
    Retrieves local affective cues such as valence, arousal, dominance, categorical emotion predictions, and emotion confidence for the target utterance and nearby context.

    \item \textbf{Interaction behavior.} 
    Retrieves turn-taking evidence such as pauses, overlap, response delays, interruption-like behavior, neighboring turns, and dialogue-level dynamics.

    \item \textbf{Speaker-level style.} 
    Retrieves speaker-relative delivery patterns, including prosody, intensity, fluency, and repeated stance-related cues across the dialogue.

    \item \textbf{Local prosodic delivery.} 
    Retrieves the target utterance's pitch, loudness, timing, and intensity cues relative to its local context and the same speaker's baseline.

    \item \textbf{Speaker comparison.} 
    Retrieves contrastive evidence comparing the two speakers' delivery, affective patterns, speaker baselines, or dialogue behavior.
\end{itemize}

\paragraph{Anchor selection policy.}
Each evidence-plan type specifies how many transcript anchors should be selected before retrieval. 
For \emph{emotion state}, \emph{speaker-level style}, and \emph{local prosodic delivery}, the system selects three to six transcript lines. 
For \emph{interaction behavior} and \emph{speaker comparison}, it selects four to six transcript lines. 
All plans require at least one \emph{primary target} line and surrounding contextual evidence. 
For speaker-comparison plans, retrieval includes contrastive evidence from both speakers, and validation marks the evidence bundle incomplete if the required speaker evidence is missing.

\paragraph{Transcript localization.}
A full-transcript locator uses the line-numbered transcript, question, and answer candidates to select relevant transcript anchors. 
The locator is constrained to identify evidence locations only. It does not choose an answer or evaluate the answer options. 
For each selected line, the system records how the line should be used in later reasoning: as the target utterance, surrounding context, same-speaker comparison, other-speaker response, or contrastive evidence.

\paragraph{Locator normalization and repair.}
Before retrieval, the system converts the locator output into a valid set of transcript anchors. 
Each selected line is converted to the internal identifier format \texttt{L\#\#\#} (for example, a numeric locator output of \texttt{15} is converted to \texttt{L015}). 
Selections that do not correspond to any transcript line are removed unless they can be recovered by matching a quoted phrase from the question. 
Duplicate line selections are merged. 

If the locator selects more lines than allowed by the evidence plan, the system keeps the most important lines first: target utterances, quoted lines from the question, immediate neighboring context, same-speaker comparison lines, and contrastive evidence. 
If the locator misses a required target line, selects too few valid lines, selects mostly invalid line identifiers, or fails to include a line explicitly quoted in the question, a repair prompt is issued. 
If repair still fails, the system selects the transcript line with the highest text similarity to the question and expands the evidence set with nearby context lines.

\paragraph{Evidence planning and retrieval.}
Given the selected anchors, the system determines which Audio Twin entries are needed for the assigned evidence-plan type. 
This step uses only the selected transcript lines rather than the full transcript, keeping later reasoning focused on localized evidence. 
Once the anchors and evidence plan are fixed, retrieval is deterministic. 
The system fetches the selected transcript turns, aligned turn cards, speaker baseline cards, and local context blocks. 
Depending on the evidence-plan type, it may also retrieve local window cards, dialogue-dynamics cards, speaker-comparison cards, representative-turn cards, or speaker-attribute records.

For each primary target line, the system constructs a local context block from nearby transcript turns in chronological order. 
The block keeps the neighboring turns, records how each selected line is used in reasoning, and summarizes whether the target occurs in a back-and-forth exchange or within a longer same-speaker stretch. 
A back-and-forth exchange is identified when nearby turns alternate repeatedly between speakers, which can indicate interactional timing such as quick responses, interruptions, or immediate reactions. 
When the nearby turns are mostly from the same speaker, the block is treated as same-speaker continuation, which is more useful for interpreting the target line relative to that speaker's surrounding speech.

\paragraph{Evidence validation.}
Before answer prediction, the system checks whether the retrieved evidence satisfies the evidence plan. 
Validation verifies that the selected anchors meet the required line count, include a primary target, contain required context, and are paired with Audio Twin entries from the correct transcript lines and speakers. 
For emotion-state, speaker-level-style, and local-prosodic-delivery plans, the required evidence includes a target transcript turn, an aligned turn card, and the corresponding speaker baseline. 
For interaction-behavior plans, it includes transcript-exchange evidence and dialogue-dynamics evidence. 
For speaker-comparison plans, it includes evidence from both speakers, a speaker-comparison card, and representative-turn cards.

If any required evidence is missing, the evidence bundle is marked incomplete and the missing evidence type is recorded. 
The pipeline may still produce a forced multiple-choice answer, but the run is flagged with incomplete retrieval and is auditable through the trace and output fields. 
Thus, selecting transcript lines alone is not treated as successful retrieval. The selected lines must be paired with the required Audio Twin evidence.

\paragraph{Prompt boundaries.}
The full line-numbered transcript is exposed only to the transcript locator. 
The evidence-planning step receives the selected lines and inspection needs, not the full transcript. 
The diagnostic-grounding step receives the validated retrieval bundle and system validation result. 
The answer stage receives the diagnostic grounding result and the set of validated evidence identifiers. 
This separation prevents later stages from introducing new transcript regions after retrieval.

\paragraph{Stage validation.}
Each model call is checked against an expected output format before its result is used by the next stage. 
The locator output must contain transcript-line selections and their intended uses, but no answer prediction. 
It is rejected if it includes answer-like fields such as a choice index, final answer, candidate comparison, or choice-level support. 
The evidence-planning output must specify retrieval needs only, and is rejected if it attempts to choose an answer or provide a text-only preference. 
The diagnostic-grounding output must include a final candidate, candidate-level evidence assessments, and an explanation of why the surface transcript reading is insufficient. 
The answer output must contain a valid zero-based choice index, and any cited evidence must come from the validated retrieval bundle.

\paragraph{Diagnostic grounding.}
The final reasoning step compares the retrieved evidence from four perspectives: the target utterance by itself, the surrounding dialogue context, the acoustic evidence retrieved from the Audio Twin, and the evidence for or against each answer candidate. 
For each candidate, the model records what would need to be true for that candidate to be correct, whether the transcript supports it, whether the local context supports it, whether the acoustic evidence supports it, and whether any evidence contradicts it. 
The choice candidate evidence is then classified as \emph{diagnostic}, \emph{compatible but not diagnostic}, \emph{contradictory}, or \emph{insufficient}.  
The final answer is selected from this diagnostic grounding result, and any cited evidence must come from the validated retrieval bundle.

\paragraph{Execution traces.}
Each run stores a JSON trace with the intermediate outputs needed to audit the pipeline: the locator output, evidence plan, retrieved evidence identifiers, retrieval validation result, evidence bundle, diagnostic-grounding output, final answer output, and validation errors. 
The tabular output records the evidence-plan type, whether the plan was satisfied, missing evidence, acoustic evidence count, retrieval status, grounding status, answer-parsing status, evidence-citation status, and the path to the trace file. 
These fields make it possible to identify whether an error came from localization, retrieval, evidence validation, grounding, answer parsing, or citation of evidence outside the validated bundle.

\section{Model Details}
\label{app:model_details}

Table~\ref{tab:model_details} lists the model names used in the paper, the corresponding checkpoints, evaluation settings, and publicly available parameter-count information. 
Reported self-hosted runs used at most 2 A100 GPUs, 8 CPUs, 64 GB RAM, and 24 h per run; hosted API runs used at most 2 local CPUs, 8 GB RAM, and 4 h per run.

\begin{table*}[t]
\centering
\scriptsize
\setlength{\tabcolsep}{3.5pt}
\renewcommand{\arraystretch}{1.15}
\begin{tabularx}{\textwidth}{@{}p{0.17\textwidth}Xccc p{0.23\textwidth}@{}}
\toprule
\textbf{Model label} & \textbf{Checkpoint} & \textbf{Text} & \textbf{Audio} & \textbf{Audio Twin} & \textbf{Public size information} \\
\midrule
Haiku 4.5; Haiku 4.5 (agentic-style)
& \nolinkurl{us.anthropic.claude-haiku-4-5-20251001-v1:0}
& Yes & No & Yes & Not publicly disclosed \\

Sonnet 4.5; Sonnet 4.5 (agentic-style)
& \nolinkurl{us.anthropic.claude-sonnet-4-5-20250929-v1:0}
& Yes & No & Yes & Not publicly disclosed \\

Kimi K2 Thinking
& \nolinkurl{moonshot.kimi-k2-thinking}
& Yes & No & No & 1T total, 32B active parameters \\

AudioFlamingoNext
& \nolinkurl{nvidia/audio-flamingo-next-hf}
& No & Yes & No & 8B parameters \\

MIMOAudio
& \nolinkurl{XiaomiMiMo/MiMo-Audio-7B-Instruct}
& Yes & Yes & No & 8B parameters \\

Qwen2.5-Omni
& \nolinkurl{Qwen/Qwen2.5-Omni-7B}
& Yes & Yes & No & 7B parameters\\

StepAudio-R1
& \nolinkurl{Step-Audio-R1}
& Yes & Yes & No & Not publicly disclosed \\

KimiAudio
& \nolinkurl{moonshotai/Kimi-Audio-7B-Instruct}
& Yes & Yes & No & 7B parameters\\

Qwen3-Omni
& \nolinkurl{Qwen/Qwen3-Omni-30B-A3B-Instruct}
& Yes & Yes & No & 30B total, 3B active parameters\\

Nova 2 Lite
& \nolinkurl{us.amazon.nova-2-lite-v1:0}
& Yes & No & No & Not publicly disclosed \\

DeepSeek V3.1
& \nolinkurl{deepseek.v3-v1:0}
& Yes & No & No & 671B total, 37B active parameters \\

StepAudio-2
& \nolinkurl{stepfun-ai/Step-Audio-2-mini}
& Yes & Yes & No & Not publicly disclosed \\

GPT-4o-Audio-Mini
& \nolinkurl{gpt-audio-mini}
& No & Yes & No & Not publicly disclosed \\
\bottomrule
\end{tabularx}
\caption{
Model checkpoints and evaluation settings used in the experiments. 
The \textbf{Model label} column shows the shorthand names used in this work, while \textbf{Checkpoint} shows the models' checkpoint identifier. 
\textbf{Text} denotes transcript-only evaluation with \((T,q,C)\), \textbf{Audio} denotes direct audio-input evaluation, and \textbf{Audio Twin} denotes evaluation with the agentic-style Audio Twin retrieval pipeline. 
Parameter counts are reported using public information when available. 
For mixture-of-experts models, total and active parameter counts are reported when available.
}
\label{tab:model_details}
\end{table*}

\section{Example Agentic Reasoning Trace}
\label{app:example_agentic_trace}

This appendix gives a compact example of one stored execution trace. The trace expands the three-stage inference procedure from Section~4.3 into the stored trace records: localization, evidence planning, retrieval validation, diagnostic grounding, and final answer output. The instance is question \texttt{i031} from conversation \texttt{V00\_S2017\_I00001189}. Choice indices are zero-based in the stored trace. The question asks:

\begin{quote}
\small
When the neighbor says ``Mm-hmm'' immediately before bringing up the noise issue, what function does this backchannel serve in the interaction?
\end{quote}

The four answer choices are: \emph{Buying time and steeling herself for the topic shift}; \emph{Enthusiastically encouraging elaboration}; \emph{Smoothly transitioning without hesitation}; and \emph{Neutrally signaling continued listening}.

\paragraph{Localization.}
The transcript locator selects evidence anchors without choosing an answer. It identifies the target backchannel, the preceding family-topic response, the following hesitation markers, the later noise complaint, and an earlier same-speaker backchannel for comparison. The stored locator output includes:

\begin{quote}
\small
``This is the 'Mm-hmm' backchannel by the neighbor (B) immediately before she shifts topic to the noise issue.''

``A's preceding turn about family that B is responding to with the backchannel; needed to understand what B is acknowledging.''

``B's 'Oh, oh, okay' follows the Mm-hmm and may indicate hesitation or topic-shift preparation.''

``The actual topic shift to the noise complaint; establishes what the backchannel was leading into.''

``Earlier multiple 'Mm-hmm' backchannels by B provide a baseline for how B typically uses this token in neutral listening contexts.''
\end{quote}

The normalized selected line identifiers are \texttt{L021}, \texttt{L025}, \texttt{L026}, \texttt{L027}, \texttt{L028}, and \texttt{L029}. The requested inspection needs are \texttt{turn\_card}, \texttt{speaker\_baseline\_card}, \texttt{local\_context\_block}, and \texttt{dialogue\_dynamics\_card}.

\paragraph{Evidence planning.}
The evidence-planning output assigns the instance to the \texttt{interaction\_behavior} plan and prevents answer selection before retrieval. The stored plan includes:

\begin{quote}
\small
``Fetch turn\_card for L026 to inspect acoustic/prosodic features of the target 'Mm-hmm' (duration, pitch contour, intensity, pause before/after, voice quality) as text-like acoustic descriptors.''

``Fetch turn\_window\_card covering L025-L029 to examine timing, inter-turn pauses, overlap, and whether L026 is followed by a notable gap before the topic-shift disfluencies in L027-L028.''

``Fetch speaker\_baseline\_card for speaker B to establish B's typical backchannel prosody (e.g., pitch, duration, energy of routine 'Mm-hmm' tokens such as L021) as a comparison baseline.''

``Compare target L026 prosodic descriptors against B's baseline backchannels (L021) to determine whether L026 deviates (e.g., longer, lower energy, trailing pause) in ways consistent with stalling/steeling vs. neutral continuer vs. encouraging vs. smooth transition.''

``Cross-reference following\_context disfluencies ('Oh, oh, okay'; 'Is, is, yeah, I'm so sorry') in L027-L028 with the timing/prosody of L026 to assess whether the backchannel functions as a pre-topic-shift hedge.''
\end{quote}

The same step records \texttt{answer\_allowed: false}. The plan also requests dialogue-level dynamics, although the final rationale in this example is driven primarily by the local context, the target turn card, and B's speaker baseline.

\paragraph{Retrieval validation.}
The retrieval stage returns the localized transcript evidence and the corresponding Audio Twin entries. For this run, validation marks the evidence contract as satisfied and permits answer selection. The trace records:

\begin{quote}
\small
\texttt{contract\_satisfied: true}

\texttt{missing\_evidence: []}

\texttt{acoustic\_evidence\_count: 9}

\texttt{text\_alone\_sufficient: false}

\texttt{answer\_allowed: true}
\end{quote}

This separates successful transcript localization from successful audio-grounded retrieval: the selected lines must be paired with the required Audio Twin evidence before the run is treated as complete.

\paragraph{Diagnostic grounding.}
The diagnostic-grounding output compares the text-only surface reading with the local dialogue context and the retrieved acoustic evidence. The isolated target is first recognized as potentially misleading:

\begin{quote}
\small
``'Mm-hmm' in isolation reads as a neutral continuer/listening token.''
\end{quote}

The local-context reading then shifts the interpretation toward choice 0:

\begin{quote}
\small
``L026 is followed by disfluent topic-shift markers: 'Oh, oh, okay' (L027), 'Is, is, yeah, I'm so sorry' (L028), before finally raising the noise complaint (L029). The backchannel sits at a hinge point where B is preparing to abandon A's family topic and introduce a face-threatening complaint, suggesting it functions as a stalling/bracing token rather than routine listenership.''
\end{quote}

The trace explicitly records the conflict between the transcript-plausible surface reading and the audio-grounded answer:

\begin{quote}
\small
``Read alone, 'Mm-hmm' looks like a generic neutral continuer (choice 3). But acoustic marking (slower, stretched, pitch-varied, lowered loudness vs. B's baseline and vs. her earlier neutral L021 backchannel) plus the immediately following disfluent restarts and topic pivot to a complaint show the token is functioning as a hesitation/bracing device before a difficult topic shift.''
\end{quote}

The diagnostic step records \texttt{text\_only\_trap\_or\_surface\_reading: 3}, \texttt{final\_candidate: 0}, and \texttt{text\_alone\_sufficient: false}.

\paragraph{Answer.}
The final answer output selects choice 0. The stored response is:

\begin{quote}
\small
``L026's 'Mm-hmm' is prosodically marked (stretched $\sim$2.46s, slower, lowered loudness, higher and more varied pitch) compared to B's baseline and her earlier neutral backchannel at L021. It is immediately followed by disfluent restarts ('Oh, oh, okay'; 'Is, is, yeah, I'm so sorry') before B introduces the face-threatening noise complaint, indicating the token functions as a stalling/bracing device rather than a neutral continuer or smooth transition.''
\end{quote}

The predicted answer is therefore:

\begin{quote}
\small
\textbf{Choice 0:} Buying time and steeling herself for the topic shift.
\end{quote}

This example shows a conflict case where the transcript-plausible answer is rejected after speaker-relative acoustic evidence and local sequential context are retrieved.

\newpage

\section{Prompt templates}
We show abbreviated versions of the main prompts used at each LLM stage. 
The full prompt templates, including complete JSON schemas, repair prompts, and validation fields, are provided in the code.

\tcbset{
  promptbox/.style={
    breakable,
    enhanced,
    sharp corners,
    colback=white,
    colframe=black,
    boxrule=\fboxrule,
    boxsep=\fboxsep,
    left=0pt,right=0pt,top=0pt,bottom=0pt,
    width=0.97\columnwidth,
    listing only,
    listing options={
      basicstyle=\scriptsize\ttfamily,
      breaklines=true,
      breakatwhitespace=false,
      columns=fullflexible,
      keepspaces=true,
      showstringspaces=false
    },
    overlay first={
      \draw[gray!45,line width=0.3pt] (frame.south west) -- (frame.south east);
    },
    overlay middle={
      \draw[gray!45,line width=0.3pt] (frame.north west) -- (frame.north east);
      \draw[gray!45,line width=0.3pt] (frame.south west) -- (frame.south east);
    },
    overlay last={
      \draw[gray!45,line width=0.3pt] (frame.north west) -- (frame.north east);
    }
  }
}

\begin{tcblisting}{
  promptbox,
  before upper={\scriptsize\bfseries Shared system prompt:\par\medskip}
}
You are an agentic multiple-choice QA assistant. Use only transcript turns and text-like acoustic evidence cards provided by tools. Never use hidden answer keys, question-generation explanations, prompt metadata, raw audio, or label files. Output strict JSON only.
\end{tcblisting}

\begin{tcblisting}{
  promptbox,
  before upper={\scriptsize\bfseries Transcript localization prompt:\par\medskip}
}
{
  "action_required": "full_transcript_locator",
  "question": "<question>",
  "choices": "<answer_candidates>",
  "line_numbered_full_transcript": "<full_transcript>",
  "schema": {
    "action": "full_transcript_locator",
    "selected_lines": [
      {
        "line_id": "L000",
        "role": "primary_target|prior_context|following_context|same_speaker_pattern|other_speaker_response|contrastive_evidence",
        "reason": "This line should be inspected because ..."
      }
    ],
    "inspection_needs": [
      "turn_card",
      "speaker_baseline_card",
      "local_context_block"
    ]
  },
  "rules": [
    "Use the full transcript only for navigation.",
    "Do not evaluate the answer options.",
    "Do not infer the final answer.",
    "Select the target lines, surrounding context, and broader lines that may need acoustic or context evidence."
  ]
}
\end{tcblisting}

\begin{tcblisting}{
  promptbox,
  before upper={\scriptsize\bfseries Evidence-planning prompt:\par\medskip}
}
{
  "action_required": "reason",
  "question": "<question>",
  "choices": "<answer_candidates>",
  "category": "<category>",
  "locator_selected_lines": "<selected_lines>",
  "locator_inspection_needs": "<inspection_needs>",
  "schema": {
    "action": "reason",
    "question_type": "lexical_content|local_prosodic_delivery|emotion_state|interaction_behavior|speaker_level_style|speaker_comparison",
    "required_evidence": ["..."],
    "optional_evidence": ["..."],
    "retrieval_plan": ["..."],
    "minimum_selected_lines": 3,
    "answer_allowed": false
  },
  "rules": [
    "Do not choose an answer.",
    "Use only the locator-selected lines and inspection needs to decide the evidence contract.",
    "Do not request or reconstruct the raw full transcript.",
    "Describe only evidence requirements and retrieval needs."
  ]
}
\end{tcblisting}

\begin{tcblisting}{
  promptbox,
  before upper={\scriptsize\bfseries Diagnostic-grounding prompt:\par\medskip}
}
{
  "action_required": "diagnostic_ground",
  "question": "<question>",
  "choices": "<answer_candidates>",
  "contract": "<evidence_plan>",
  "retrieved_evidence_bundle": "<validated_evidence>",
  "system_validation": "<validation_result>",
  "schema": {
    "action": "diagnostic_ground",
    "isolated_target_reading": "<target_line_reading>",
    "local_context_reading": "<context_reading>",
    "acoustic_reading": "<audio_twin_reading>",
    "choice_tests": [
      {
        "choice_index": 0,
        "required_condition": "...",
        "text_support": "diagnostic|compatible_but_not_diagnostic|contradictory|insufficient",
        "local_context_support": "diagnostic|compatible_but_not_diagnostic|contradictory|insufficient",
        "acoustic_support": "diagnostic|compatible_but_not_diagnostic|contradictory|insufficient",
        "diagnostic_status": "diagnostically_supported|not_diagnostic|contradicted|insufficient"
      }
    ],
    "final_candidate": 0,
    "text_alone_sufficient": false
  },
  "rules": [
    "Compare isolated target-line, surrounding-context, acoustic-card, and choice-level diagnostic interpretations.",
    "Do not treat compatibility as diagnostic support.",
    "Intent and stance are primarily established by dialogue context; delivery features support, weaken, or qualify that interpretation.",
    "Treat locator-selected lines as navigation anchors; answer reasoning must come from retrieved evidence cards and local context blocks."
  ]
}
\end{tcblisting}

\begin{tcblisting}{
  promptbox,
  before upper={\scriptsize\bfseries Answer prompt:\par\medskip}
}
{
  "action_required": "answer",
  "question": "<question>",
  "choices": "<answer_candidates>",
  "grounding_result": "<diagnostic_grounding>",
  "validated_evidence_ids": "<validated_evidence_ids>",
  "schema": {
    "action": "answer",
    "choice_index": 0,
    "choice_text": "exact choice text",
    "reason": "concise evidence-based reason",
    "evidence_used": ["..."],
    "text_alone_sufficient": false
  },
  "rules": [
    "Choose only from the diagnostic grounding result.",
    "Do not introduce new facts, retrieval candidates, raw transcript, raw features, or evidence.",
    "choice_index must be 0, 1, 2, or 3."
  ]
}
\end{tcblisting}

\bibliography{neurips_2026}

\begin{thebibliography}{23}
\providecommand{\natexlab}[1]{#1}
\providecommand{\url}[1]{\texttt{#1}}
\expandafter\ifx\csname urlstyle\endcsname\relax
  \providecommand{\doi}[1]{doi: #1}\else
  \providecommand{\doi}{doi: \begingroup \urlstyle{rm}\Url}\fi

\bibitem[Agrawal et~al.(2025)Agrawal, Akinyemi, Alvero, Behrooz, Buffalini, Carlucci, Chen, Chen, Chen, Cheng, et~al.]{agrawal2025seamless}
Vasu Agrawal, Akinniyi Akinyemi, Kathryn Alvero, Morteza Behrooz, Julia Buffalini, Fabio~Maria Carlucci, Joy Chen, Junming Chen, Zhang Chen, Shiyang Cheng, et~al.
\newblock Seamless interaction: Dyadic audiovisual motion modeling and large-scale dataset.
\newblock \emph{arXiv preprint arXiv:2506.22554}, 2025.

\bibitem[Castro et~al.(2019)Castro, Hazarika, P{\'e}rez-Rosas, Zimmermann, Mihalcea, and Poria]{castro2019towards}
Santiago Castro, Devamanyu Hazarika, Ver{\'o}nica P{\'e}rez-Rosas, Roger Zimmermann, Rada Mihalcea, and Soujanya Poria.
\newblock Towards multimodal sarcasm detection (an \_obviously\_ perfect paper).
\newblock In \emph{Proceedings of the 57th annual meeting of the association for computational linguistics}, pages 4619--4629, 2019.

\bibitem[Chu et~al.(2023)Chu, Xu, Zhou, Yang, Zhang, Yan, Zhou, and Zhou]{chu2023qwen}
Yunfei Chu, Jin Xu, Xiaohuan Zhou, Qian Yang, Shiliang Zhang, Zhijie Yan, Chang Zhou, and Jingren Zhou.
\newblock Qwen-audio: Advancing universal audio understanding via unified large-scale audio-language models.
\newblock \emph{arXiv preprint arXiv:2311.07919}, 2023.

\bibitem[Feng et~al.(2025)Feng, Lee, Xu, Lee, Lertpetchpun, Shi, Wang, Thebaud, Moro-Velazquez, Byrd, et~al.]{feng2025vox}
Tiantian Feng, Jihwan Lee, Anfeng Xu, Yoonjeong Lee, Thanathai Lertpetchpun, Xuan Shi, Helin Wang, Thomas Thebaud, Laureano Moro-Velazquez, Dani Byrd, et~al.
\newblock Vox-profile: A speech foundation model benchmark for characterizing diverse speaker and speech traits.
\newblock \emph{arXiv preprint arXiv:2505.14648}, 2025.

\bibitem[Geirhos et~al.(2020)Geirhos, Jacobsen, Michaelis, Zemel, Brendel, Bethge, and Wichmann]{geirhos2020shortcut}
Robert Geirhos, J{\"o}rn-Henrik Jacobsen, Claudio Michaelis, Richard Zemel, Wieland Brendel, Matthias Bethge, and Felix~A Wichmann.
\newblock Shortcut learning in deep neural networks.
\newblock \emph{Nature Machine Intelligence}, 2\penalty0 (11):\penalty0 665--673, 2020.

\bibitem[Goyal et~al.(2017)Goyal, Khot, Summers-Stay, Batra, and Parikh]{goyal2017making}
Yash Goyal, Tejas Khot, Douglas Summers-Stay, Dhruv Batra, and Devi Parikh.
\newblock Making the {V} in {VQA} matter: Elevating the role of image understanding in {Visual Question Answering}.
\newblock In \emph{Proceedings of the IEEE conference on computer vision and pattern recognition}, pages 6904--6913, 2017.

\bibitem[Huang et~al.(2024{\natexlab{a}})Huang, Lu, Wang, Hsiao, Kuan, Wu, Arora, Chang, Shi, Peng, et~al.]{huang2024dynamic}
Chien-yu Huang, Ke-Han Lu, Shih-Heng Wang, Chi-Yuan Hsiao, Chun-Yi Kuan, Haibin Wu, Siddhant Arora, Kai-Wei Chang, Jiatong Shi, Yifan Peng, et~al.
\newblock Dynamic-superb: Towards a dynamic, collaborative, and comprehensive instruction-tuning benchmark for speech.
\newblock In \emph{ICASSP 2024-2024 IEEE International Conference on Acoustics, Speech and Signal Processing (ICASSP)}, pages 12136--12140. IEEE, 2024{\natexlab{a}}.

\bibitem[Huang et~al.(2024{\natexlab{b}})Huang, Li, Yang, Shi, Chang, Ye, Wu, Hong, Huang, Liu, et~al.]{huang2024audiogpt}
Rongjie Huang, Mingze Li, Dongchao Yang, Jiatong Shi, Xuankai Chang, Zhenhui Ye, Yuning Wu, Zhiqing Hong, Jiawei Huang, Jinglin Liu, et~al.
\newblock Audiogpt: Understanding and generating speech, music, sound, and talking head.
\newblock In \emph{Proceedings of the AAAI conference on artificial intelligence}, volume~38, pages 23802--23804, 2024{\natexlab{b}}.

\bibitem[Jadoul et~al.(2018)Jadoul, Thompson, and De~Boer]{jadoul2018introducing}
Yannick Jadoul, Bill Thompson, and Bart De~Boer.
\newblock Introducing parselmouth: A python interface to praat.
\newblock \emph{Journal of Phonetics}, 71:\penalty0 1--15, 2018.

\bibitem[Kiela et~al.(2020)Kiela, Firooz, Mohan, Goswami, Singh, Ringshia, and Testuggine]{kiela2020hateful}
Douwe Kiela, Hamed Firooz, Aravind Mohan, Vedanuj Goswami, Amanpreet Singh, Pratik Ringshia, and Davide Testuggine.
\newblock The hateful memes challenge: Detecting hate speech in multimodal memes.
\newblock In \emph{Advances in Neural Information Processing Systems}, volume~33, pages 2611--2624, 2020.

\bibitem[Poria et~al.(2019)Poria, Majumder, Mihalcea, and Hovy]{poria2019emotion}
Soujanya Poria, Navonil Majumder, Rada Mihalcea, and Eduard Hovy.
\newblock Emotion recognition in conversation: Research challenges, datasets, and recent advances.
\newblock \emph{IEEE Access}, 7:\penalty0 100943--100953, 2019.

\bibitem[Radford et~al.(2023)Radford, Kim, Xu, Brockman, McLeavey, and Sutskever]{radford2023robust}
Alec Radford, Jong~Wook Kim, Tao Xu, Greg Brockman, Christine McLeavey, and Ilya Sutskever.
\newblock Robust speech recognition via large-scale weak supervision.
\newblock In \emph{International conference on machine learning}, pages 28492--28518. PMLR, 2023.

\bibitem[Sakshi et~al.(2025)Sakshi, Tyagi, Kumar, Seth, Selvakumar, Nieto, Duraiswami, Ghosh, and Manocha]{sakshi2025mmau}
Sakshi Sakshi, Utkarsh Tyagi, Sonal Kumar, Ashish Seth, Ramaneswaran Selvakumar, Oriol Nieto, Ramani Duraiswami, Sreyan Ghosh, and Dinesh Manocha.
\newblock Mmau: A massive multi-task audio understanding and reasoning benchmark.
\newblock In \emph{International Conference on Learning Representations}, volume 2025, pages 84929--84964, 2025.

\bibitem[Schick et~al.(2023)Schick, Dwivedi-Yu, Dess{\`\i}, Raileanu, Lomeli, Hambro, Zettlemoyer, Cancedda, and Scialom]{schick2023toolformer}
Timo Schick, Jane Dwivedi-Yu, Roberto Dess{\`\i}, Roberta Raileanu, Maria Lomeli, Eric Hambro, Luke Zettlemoyer, Nicola Cancedda, and Thomas Scialom.
\newblock Toolformer: Language models can teach themselves to use tools.
\newblock \emph{Advances in neural information processing systems}, 36:\penalty0 68539--68551, 2023.

\bibitem[Shen et~al.(2025)Shen, Li, Liu, Li, Porras, and Unberath]{shen2025operating}
Yiqing Shen, Chenjia Li, Bohan Liu, Cheng-Yi Li, Tito Porras, and Mathias Unberath.
\newblock Operating room workflow analysis via reasoning segmentation over digital twins.
\newblock In \emph{International Conference on Medical Image Computing and Computer-Assisted Intervention}, pages 415--424. Springer, 2025.

\bibitem[Shen et~al.(2023)Shen, Song, Tan, Li, Lu, and Zhuang]{shen2023hugginggpt}
Yongliang Shen, Kaitao Song, Xu~Tan, Dongsheng Li, Weiming Lu, and Yueting Zhuang.
\newblock Hugginggpt: Solving ai tasks with chatgpt and its friends in hugging face.
\newblock \emph{Advances in Neural Information Processing Systems}, 36:\penalty0 38154--38180, 2023.

\bibitem[Tang et~al.(2024)Tang, Yu, Sun, Chen, Tan, Li, Lu, Ma, and Zhang]{tang2024salmonn}
Changli Tang, Wenyi Yu, Guangzhi Sun, Xianzhao Chen, Tian Tan, Wei Li, Lu~Lu, Zejun Ma, and Chao Zhang.
\newblock {SALMONN}: Towards generic hearing abilities for large language models.
\newblock In \emph{The Twelfth International Conference on Learning Representations}, 2024.

\bibitem[Wagner et~al.(2023)Wagner, Triantafyllopoulos, Wierstorf, Schmitt, Burkhardt, Eyben, and Schuller]{wagner2023dawn}
Johannes Wagner, Andreas Triantafyllopoulos, Hagen Wierstorf, Maximilian Schmitt, Felix Burkhardt, Florian Eyben, and Bj{\"o}rn~W Schuller.
\newblock Dawn of the transformer era in speech emotion recognition: closing the valence gap.
\newblock \emph{IEEE Transactions on Pattern Analysis and Machine Intelligence}, 45\penalty0 (9):\penalty0 10745--10759, 2023.

\bibitem[Wang et~al.(2025)Wang, Zou, Lin, Sun, Liu, Zhang, Liu, Aw, and Chen]{wang2025audiobench}
Bin Wang, Xunlong Zou, Geyu Lin, Shuo Sun, Zhuohan Liu, Wenyu Zhang, Zhengyuan Liu, AiTi Aw, and Nancy Chen.
\newblock Audiobench: A universal benchmark for audio large language models.
\newblock In \emph{Proceedings of the 2025 Conference of the Nations of the Americas Chapter of the Association for Computational Linguistics: Human Language Technologies (Volume 1: Long Papers)}, pages 4297--4316, 2025.

\bibitem[Wang et~al.(2020)Wang, Tran, and Feiszli]{wang2020makes}
Weiyao Wang, Du~Tran, and Matt Feiszli.
\newblock What makes training multi-modal classification networks hard?
\newblock In \emph{Proceedings of the IEEE/CVF conference on computer vision and pattern recognition}, pages 12695--12705, 2020.

\bibitem[Wu et~al.(2022)Wu, Jastrzebski, Cho, and Geras]{wu2022characterizing}
Nan Wu, Stanislaw Jastrzebski, Kyunghyun Cho, and Krzysztof~J Geras.
\newblock Characterizing and overcoming the greedy nature of learning in multi-modal deep neural networks.
\newblock In \emph{International Conference on Machine Learning}, pages 24043--24055. PMLR, 2022.

\bibitem[Yang et~al.(2024)Yang, Xu, Liu, Chu, Jiang, Zhou, Leng, Lv, Zhao, Zhou, et~al.]{yang2024air}
Qian Yang, Jin Xu, Wenrui Liu, Yunfei Chu, Ziyue Jiang, Xiaohuan Zhou, Yichong Leng, Yuanjun Lv, Zhou Zhao, Chang Zhou, et~al.
\newblock Air-bench: Benchmarking large audio-language models via generative comprehension.
\newblock In \emph{Proceedings of the 62nd Annual Meeting of the Association for Computational Linguistics (Volume 1: Long Papers)}, pages 1979--1998, 2024.

\bibitem[Yao et~al.(2023)Yao, Zhao, Yu, Du, Shafran, Narasimhan, and Cao]{yao2023react}
Shunyu Yao, Jeffrey Zhao, Dian Yu, Nan Du, Izhak Shafran, Karthik Narasimhan, and Yuan Cao.
\newblock {ReAct}: Synergizing reasoning and acting in language models.
\newblock In \emph{The Eleventh International Conference on Learning Representations}, 2023.

\end{thebibliography}
\bibliographystyle{plainnat}

\end{document}